\documentclass[10pt,twocolumn,letterpaper]{article}
\usepackage[T1]{fontenc}
\usepackage[pagenumbers]{cvpr}
\usepackage{placeins}
\usepackage{array,multirow}
\usepackage{cuted}
\usepackage{currfile}
\definecolor{cvprblue}{rgb}{0.21,0.49,0.74}
\usepackage[pagebackref,breaklinks,colorlinks,allcolors=cvprblue]{hyperref}
\makeatletter
\patchcmd{\NAT@citexnum}
  {\let\NAT@last@num\NAT@num}
  {\Hy@backout{\@citeb\@extra@b@citeb}\let\NAT@last@num\NAT@num}
  {}{\PackageError{RoboField}{Could not repair citation page backreferences}{}}
\makeatother
\hypersetup{pdfauthor={},pdftitle={RoboField: Unstructured Scene Generation and Task-Adaptive Simulation}}
\def\confName{CVPR}
\def\confYear{2027}

\usepackage[nospread,noshrink,nocheckfloat,nocheckfootnote]{flushend}
\makeatletter
\patchcmd{\@maketitle}{\vskip .375in}{\vskip .055in}{}{}
\patchcmd{\@maketitle}{\vspace*{24pt}}{\vspace*{3pt}}{}{}
\patchcmd{\@maketitle}{\vskip .5em}{\vskip 0pt}{}{}
\patchcmd{\@maketitle}{\vspace*{12pt}}{\vspace*{1pt}}{}{}
\makeatother
\newcommand*{\RoboFieldTitle}{RoboField: Unstructured Scene Generation and Task-Adaptive Simulation}
\newcommand{\RoboFieldAuthors}{%
  Yujia Xie\textsuperscript{1,2,*}\quad Pengjun Niu\textsuperscript{1,2,*}\quad
  Yimo Wang\textsuperscript{1,2}\quad Yubo Zhao\textsuperscript{3}\quad
  Lechen Ruan\textsuperscript{4}\quad Ke Liu\textsuperscript{1,2,\textdagger}\\[2pt]
  \normalsize\textsuperscript{1}School of Advanced Manufacturing and Robotics, Peking University, Beijing, 100871, China\\
  \normalsize\textsuperscript{2}Research Center for Robotics, Peking University, Beijing, 100871, China\\
  \normalsize\textsuperscript{3}School of Information Engineering, Chang'an University, Xi'an 710064, China\qquad
  \textsuperscript{4}HEC Future Lab\\
  \small\textsuperscript{*}Equal contribution.\quad
  \textsuperscript{\textdagger}Corresponding author
  (\href{mailto:liuke@pku.edu.cn}{\nolinkurl{liuke@pku.edu.cn}}).%
}
\title{\RoboFieldTitle}
\author{\RoboFieldAuthors}
\hypersetup{pdfauthor={Yujia Xie, Pengjun Niu, Yimo Wang, Yubo Zhao, Lechen Ruan, Ke Liu}}
\begin{document}
\maketitle
\begin{strip}
\centering\includegraphics[width=0.87\textwidth]{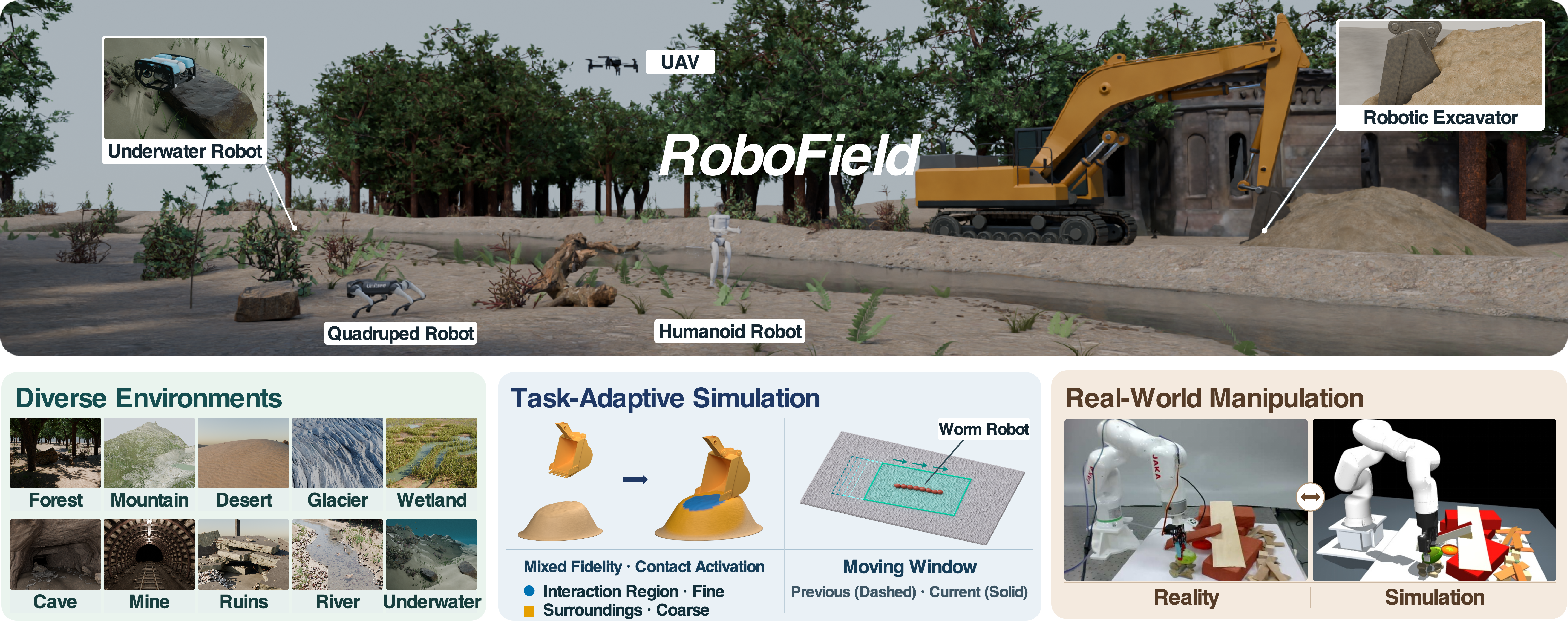}
\captionsetup{hypcap=false}
\captionof{figure}{\textbf{RoboField for unstructured environments.} A shared scene representation connects scene generation with task-adaptive simulation to support diverse robots and tasks.}
\captionsetup{hypcap=true}
\label{fig:overview}
\end{strip}

\begingroup
\renewcommand{\endabstract}{\par}
\begin{abstract}
Interactive unstructured scenes are essential for robotics, yet their generation and physical simulation remain particularly time-consuming. RoboField connects multimodal-model-guided scene planning, retrieval and on-demand generation of missing assets, and procedural synthesis to task-adaptive simulation through a shared scene representation. We construct ten environment categories with five visual configurations and use the generated scenes for robot interaction, navigation planning evaluation, and reinforcement learning-based navigation and motion-control training. We further assess task-adaptive simulation through excavation, worm locomotion, and real-robot manipulation replay. Compared with a uniformly fine numerical reference, excavation runtime decreases by 22.8\%, with a 0.79\% final bucket-load difference. Compared with full-domain simulation, worm locomotion runtime decreases by 44.0\%, with a 3D center-of-mass trajectory root-mean-square error (RMSE) of 0.338 mm. Across three layouts and 64 recorded JAKA Mini2 trajectories, task-adaptive replay reduces runtime by 15.0\%, while simulated task outcomes agree with real-world observations in 98.44\% of cases.
\end{abstract}

\endgroup
{\small\urlstyle{same}\noindent \textbf{Project page:} \url{http://www.liukepku.com/robofield/index.html}\par}
\begin{figure*}[t]
\centering{\setlength{\fboxsep}{0pt}\colorbox{white}{\includegraphics[width=\textwidth]{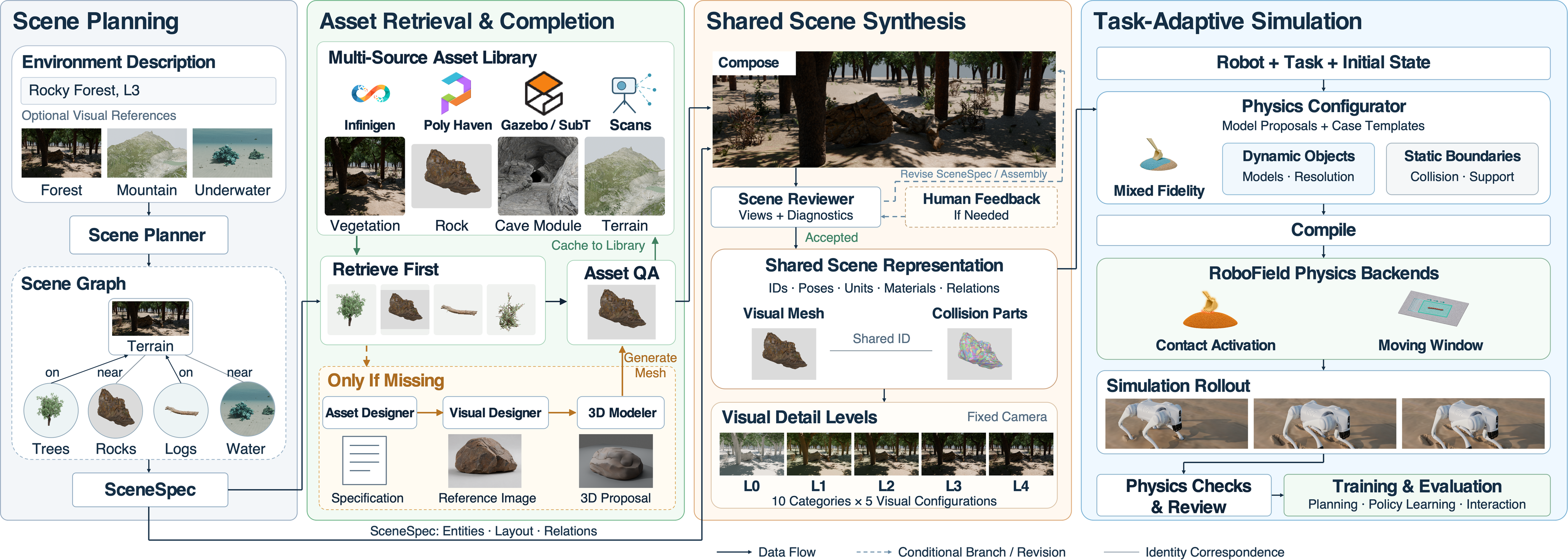}}}
\caption{\textbf{Overview of the RoboField framework.} Scene planning, asset retrieval and on-demand generation, and iterative review produce a shared scene. The Physics Configurator combines robot and task information with case templates to assign dynamic entities, static boundaries, and physical representations through mixed-fidelity modeling. RoboField Physics Backends execute the configured models with contact activation and moving windows with archived states while retaining persistent entity identities.}\label{fig:framework}
\end{figure*}

\section{Introduction}
Generating diverse, interactive robot simulation environments is essential for developing robots that perceive, traverse, and manipulate the physical world. These environments support algorithm development and evaluation. They enable repeatable experiments at lower cost than repeatedly building and testing in real environments~\cite{infinigen,procthor,habitat2019}. In simulation, contacts, support relationships, and material responses allow robots to interact with generated scenes~\cite{physcene,behavior1k2023}. This capability is particularly important in unstructured environments. Terrain, vegetation, rubble, and water affect both perception and action. Traversing uneven ground, clearing obstacles, or deforming a granular bed requires the surrounding scene to participate physically in the task. Interactive scene generation must therefore connect diverse visual content to physical representations and reproducible interaction states.

Natural scene generators can provide rich geometry and appearance~\cite{infinigen}. Language and generative models can extend layout and content creation~\cite{layoutgpt,text2room,holodeck}. Current research on interactive simulation scenes mainly focuses on object manipulation in indoor environments~\cite{physcene,igibson2022,maniskill2023,robocasa}. Outdoor scene generation often emphasizes visual appearance without modeling robot interaction. Beyond indoor object manipulation, unstructured environments involve contact and deformation of surrounding ground and materials over large areas. These interactions require suitable collision geometry and material models, whose fine resolution can make even local tasks expensive.

A visually complete environment does not require every visible entity to be dynamically simulated at the same level of detail. Relevant objects, material models, and active regions can change with the robot and task even when the underlying scene is unchanged. This observation exposes three coupled problems. Visual assets and physical models need persistent correspondence as a scene changes. Each task needs the objects and material domains that can affect its outcome, while the rest of the scene still provides visual and collision context. Local activity can move with the robot, so material state must persist when it leaves and later reenters the computed region.

RoboField addresses these problems by connecting scene generation with task-adaptive simulation through a shared scene representation (Figs.~\ref{fig:overview} and~\ref{fig:framework}). Scene planning, asset retrieval and on-demand generation, and shared scene synthesis produce this representation. The Physics Configurator selects task-relevant entities and assigns physical representations, while RoboField Physics Backends execute the configured models using contact activation and moving windows with archived states.

We make three contributions:

\noindent\textbf{(i) Shared scene synthesis and representation.} RoboField combines multimodal-model-guided scene planning, asset retrieval and on-demand generation, and procedural synthesis to construct unstructured environments. Persistent entity identities, geometry, and spatial relations provide a shared representation for task-specific physical compilation.

\noindent\textbf{(ii) Task-adaptive simulation.} A multimodal-model-assisted Physics Configurator combines task requirements, case templates, and programmatic checks to select relevant entities and assign physical representations. Mixed-fidelity modeling, contact activation, and moving windows with archived states concentrate computation on task-relevant interactions while preserving static collision and support context.

\noindent\textbf{(iii) Evaluation across planning, learning, and interaction.} We evaluate generated environments through robot-specific navigation planning and reinforcement learning-based training of navigation and motion-control policies. Excavation, locomotion, and paired replay of 64 measured robot trajectories further assess computational efficiency and agreement with numerical and real-world references.

\begin{figure*}[t]
\centering
{\setlength{\fboxsep}{0pt}\colorbox{white}{\includegraphics[width=0.82\textwidth]{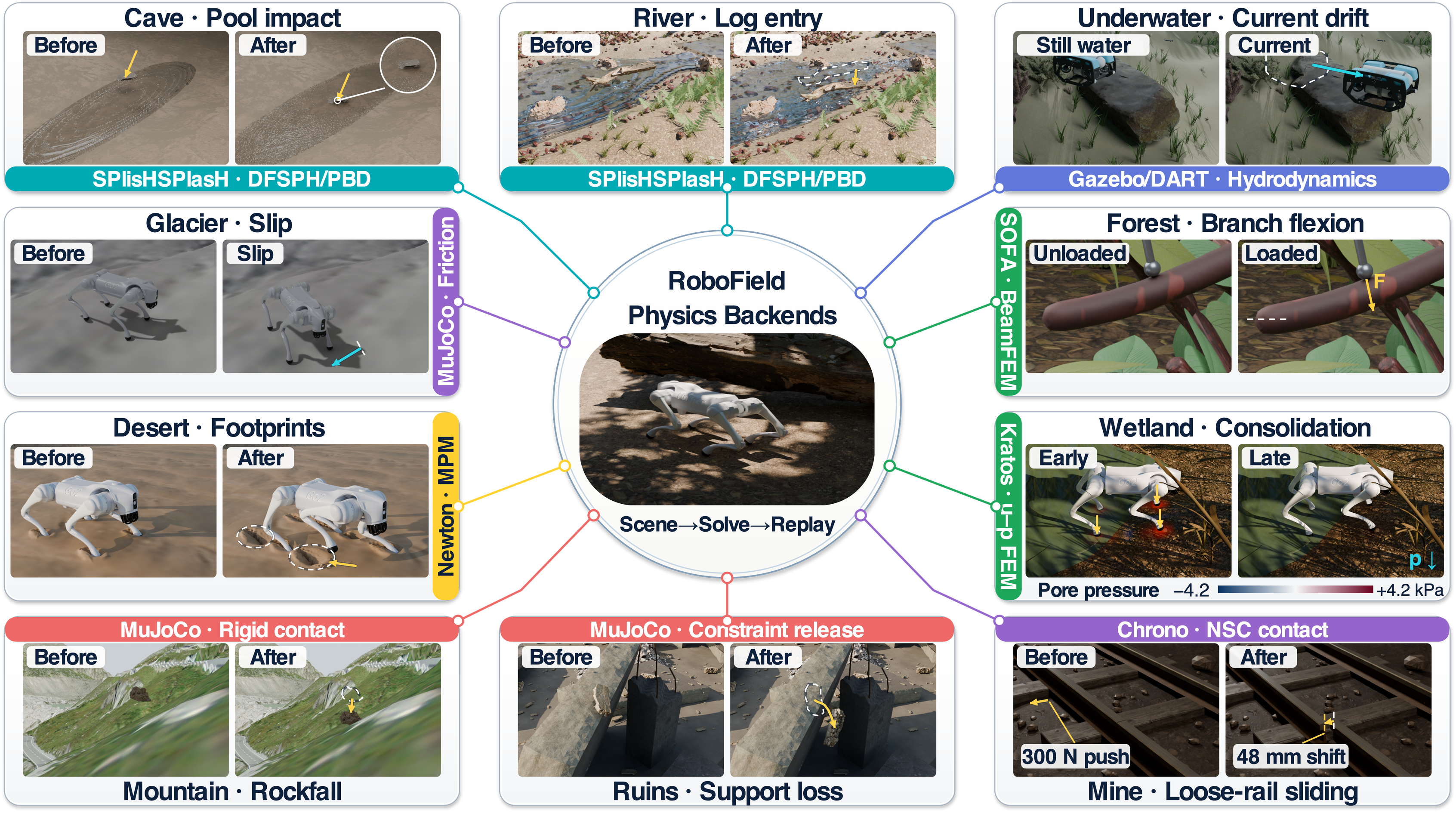}}}
\caption{\textbf{Physical interactions in generated environments.} Examples span rigid contact, terrain deformation, fluid interaction, and structural response. Each panel pairs two interaction states and identifies the physical backend and numerical model used for that example. Backend abbreviations are defined in Supplementary Sec.~B.}\label{fig:physics}
\end{figure*}

\section{Related work}
\noindent\textbf{Scene generation.}
Visual scene generation combines reusable asset collections~\cite{threedfront,threedfuture,objaverse,objaversexl} with procedural or learned models. Infinigen constructs diverse natural environments procedurally~\cite{infinigen}. Learned object arrangements~\cite{sceneformer,atiss,diffuscene} and language or image guidance~\cite{layoutgpt,text2room} extend control over layouts and appearance. Outdoor synthetic datasets vary seasons, illumination, fog, rain, and viewpoint~\cite{ros2016synthia,cabon2020virtualkitti2}. These approaches primarily address visual content and viewing conditions. RoboField addresses visual scene generation as well as the subsequent allocation of physical representations and dynamic computation for robot interaction.

\noindent\textbf{Robot simulation environments.}
Robot simulation environments and benchmarks combine physics engines~\cite{mujoco,isaacgym2021} with scene and task design for navigation~\cite{gibson2018,habitat2019}, household manipulation~\cite{igibson2022,behavior1k2023,sapien2020,maniskill2023}, and evaluation across different robot embodiments~\cite{niu2026skelwam}. ProcTHOR, Holodeck, and Infinigen Indoors expand interactive indoor generation through procedural construction and language guidance~\cite{procthor,holodeck,infinigen_indoors}. Physical and task constraints further shape generation. PhyScene incorporates physical plausibility and interaction constraints into layouts~\cite{physcene}, while RoboGen jointly generates environments, tasks, and supervision~\cite{robogen}. RoboCasa broadens household task diversity~\cite{robocasa}, and SAGE uses semantic, visual, and physical critics for iterative scene revision~\cite{sage2026}. Most of these systems emphasize indoor interaction. Meanwhile, locomotion curricula vary the difficulty of slopes, rough terrain, stairs, and obstacles according to robot performance~\cite{rudin2022learningtowalk}. RoboField connects diverse unstructured scenes to task conditioned physical representations and selective computation. Its backend adapters bind relevant entities to rigid, granular, fluid, or deformable models, supporting terrain and material interaction while retaining static collision and support context.

\noindent\textbf{Adaptive physical computation.}
Position-based dynamics (PBD), extended PBD (XPBD), the material point method (MPM), and affine particle-in-cell (APIC) methods define material response and state updates~\cite{muller2007pbd,xpbd,stomakhin2013snow,jiang2015apic}. Computational allocation complements these models. Adaptive particle sampling varies spatial resolution~\cite{adams2007adaptive}, and adaptive articulated dynamics selects degrees of freedom (DoF) using motion error~\cite{redon2005}. MuJoCo's sleeping mechanism suspends stationary constraint islands and wakes them upon contact with active bodies~\cite{mujoco_sleeping340}. RoboField organizes allocation around robot tasks. Mixed-fidelity modeling adapts physical representations and resolution to task requirements, while contact activation and moving windows control when and where dynamic simulation occurs. Together, these mechanisms focus computation on task-relevant interactions.

\section{Method}
\begin{figure*}[t]
\centering
\includegraphics[width=\textwidth]{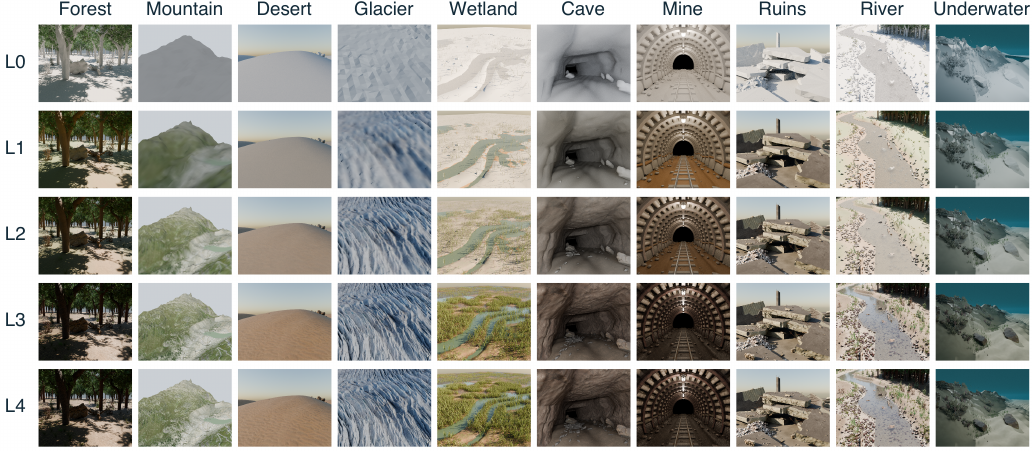}
\caption{\textbf{Scene diversity and visual detail control.} Columns show ten environment categories, and rows show visual configurations from L0 to L4. Each column presents variants of the same base scene from a fixed camera view, illustrating changes in geometry, materials, and decoration.}\label{fig:scene-matrix}
\end{figure*}

\subsection{Shared scene representation and synthesis}
RoboField produces both a renderable environment and a structured scene record for downstream task compilation. Every entity receives a stable identifier linked to its semantic type, source, pose, visual and collision geometry, mobility, material fields, and spatial relations. Validators check references, units, and poses. The inputs are an environment description $d$, optional visual references $v$ that guide scene appearance, a spatial extent $e$, a visual configuration $\ell$, and a random seed $s$. Scene generation produces a complete environment from these inputs. We denote the shared structured record by $\mathcal S$ and the generation mapping by $\mathcal G$:
\begin{equation}
\label{eq:scene-synthesis}
\mathcal S=\mathcal G(d,v,e,\ell,s).
\end{equation}

The Scene Planner and Scene Reviewer use pretrained multimodal models. The planner produces SceneSpec, a structured specification of entities, asset requirements, and spatial relations that guides asset retrieval and scene assembly, while the reviewer evaluates rendered views against scene requirements.

Asset retrieval is attempted first. On-demand generation is used when no suitable asset is found: the Asset Designer formulates its specification, the Visual Designer generates a reference image, and the 3D Modeler produces a candidate mesh. Procedural construction can also supply candidates. Asset quality assurance (Asset QA) checks semantic suitability, scale, geometry, material dependencies, and provenance before candidates are admitted to scene assembly. Environment specific Blender/Python programs construct the scene. The Scene Reviewer compares renders with the request and checks missing dependencies, intersections, disconnected geometry, and invalid placement. The loop ends on acceptance. Persistent failures go to human revision or rejection.

L0--L4 denote increasing levels of visual detail (see Supplementary Sec.~A). The shared scene representation links entity identities and poses to collision and material domains, while semantic and spatial relations support task-specific physics compilation.

\subsection{Task-conditioned physics compilation}
\noindent\textbf{Mixed-fidelity modeling.}
RoboField allocates physical detail at both object and spatial levels. At the object level, task-relevant movable objects and supports retain dynamic responses, while other geometry serves as static collision and support boundaries. At the spatial level, particle-based domains use finer resolution near anticipated interactions and coarser resolution elsewhere.

The Physics Configurator consumes the scene record $\mathcal S$, robot $r$, task $\tau$, and initial state $q_0$. When available, prescribed or recorded robot motion also informs entity selection. The multimodal model proposes entity identifiers and representation categories. Case code resolves them, follows support relations, and adds geometry reached by the robot or tool. A case template supplies solver, material, discretization, activation, and window parameters. The resulting configuration is
\begin{equation}
\label{eq:physics-configuration}
\Gamma=\mathcal F_{\mathrm{cfg}}(\mathcal S,r,\tau,q_0,\mathcal M,\mathcal T)
=\left(\mathcal I,\{\beta_i,h_i\}_{i\in\mathcal I},\omega\right),
\end{equation}
where $\mathcal M$ is the pretrained multimodal model and $\mathcal T$ is the case template. The set $\mathcal I$ contains task relevant objects or material domains. Each $\beta_i$ specifies a representation and backend binding, $h_i$ sets its discretization, and $\omega$ stores window bounds and event thresholds.

The set $\mathcal I$ is fixed before the task or measured trajectory begins. Runtime rules change activity within it. Unselected entities remain static colliders until a new compilation promotes them. RoboField stores visual presence, dynamic capability, and integration activity separately.

Backend adapters map the shared scene record and task configuration to backend-specific dynamic states and static boundaries:
\begin{equation}
\label{eq:backend-compilation}
\mathcal P=\mathcal D(\mathcal S,r,\Gamma)
=\left(\mathcal P_{\mathrm{dyn}},\mathcal B\right),
\end{equation}
where $\mathcal P_{\mathrm{dyn}}$ is the dynamic solver state and $\mathcal B$ is static collision and support geometry. The adapters preserve entity identities, transforms, material attributes, and spatial relations. Task-relevant entities are compiled into rigid, granular, fluid, or deformable models (Fig.~\ref{fig:physics}), while other geometry provides static boundaries. The shared scene representation also supports robot-specific planning abstractions and feasibility constraints. Discretization is determined during compilation, while activation and state-management policies govern runtime execution.

For particle-based physical representations, RoboField identifies a locally refined region $\Omega_f$ according to the robot task and its anticipated interaction extent. Mixed-resolution discretization concentrates physical detail on task-relevant material regions. Parent cell $p$ is assigned target diameter $h_p$ from its initial position $\mathbf x_p^0$, using fine and coarse particle diameters $h_f$ and $h_c$:
\begin{equation}
\label{eq:mixed-resolution}
h_p=
\begin{cases}
h_f,&\mathbf x_p^0\in\Omega_f,\\
h_c,&\text{otherwise},
\end{cases}
\qquad h_f<h_c.
\end{equation}
Parent cells within the refined region are subdivided into child particles before simulation. Their positions are adjusted to satisfy domain and neighboring-particle constraints while preserving total mass:
\begin{equation}
\label{eq:mass-preservation}
\sum_{j=1}^{K_p}m_{p,j}=m_p,
\end{equation}
where $m_p$ is the parent-cell mass, $m_{p,j}$ is the mass of child $j$, and $K_p$ is the number of children. Particles at different resolutions participate in a shared neighborhood search and physical solver. The discretization remains fixed during simulation, without online particle splitting or merging.

Physical compilation defines the available representations, while contact activation and moving windows with archived states control when and where dynamic integration occurs.

\subsection{Selective execution and persistent state}
\noindent\textbf{Contact activation.}
Contact activation controls when a configured particle domain begins dynamic integration. For group $g$, initialize $a_g^0=0$. At physical substep $u\geq1$, update the binary flag $a_g^u\in\{0,1\}$ using its starting robot state $q_u$:
\begin{equation}
\label{eq:contact-activation}
a_g^u=a_g^{u-1}\lor
\mathbf 1\!\left[
\operatorname{dist}\!\left(\mathcal C_r(q_u),\Omega_g\right)\leq\epsilon_g
\right],
\end{equation}
where $\mathbf 1[\cdot]$ is the indicator function, $\mathcal C_r(q_u)$ is the robot or tool collision surface, and $\Omega_g$ is the union of particle spheres. The distance is the minimum clearance after backend contact-gap and collision-margin adjustments; $\epsilon_g$ is the configured predictive margin. A broad phase and convex plane rejection precede exact point to triangle distance queries. The updated flag $a_g^u$ immediately controls that substep's particle solve. The first positive result activates the complete pile at both resolutions, and the pile remains dynamic for the rest of the run.

\noindent\textbf{Moving windows with archived states.}
The moving window restricts dynamic integration to a fixed size region. At control sample $n\geq1$ after each interval, $\mathbf p_n$ and $\mathbf w_n$ denote the robot center of mass (COM) and window center. Initialize $k_0=0$, $\mathbf p_0$ as the initial COM, and $\mathbf w_0$ as the configured horizontal center. For shift distance $\delta>0$:
\begin{equation}
\label{eq:moving-window}
\begin{aligned}
k_n&=\max\!\left(k_{n-1},\left\lfloor\frac{p_{n,x}-p_{0,x}}{\delta}\right\rfloor\right),\\
\mathbf w_n&=\mathbf w_0+k_n\delta\mathbf e_x,
\qquad W_n=\mathbf w_n+\mathcal W_{\omega}.
\end{aligned}
\end{equation}
Here $k_n$ counts forward shifts, $\mathbf e_x$ is the forward unit vector, and $W_n$ is the world-coordinate window. The template $\mathcal W_{\omega}$ is the fixed XY rectangle centered horizontally at the origin and extended along $z$. The window changes only when the center of mass crosses a successive forward threshold. A robot margin protects nearby particles. Static support tiles and a halo provide local boundary conditions.

\noindent\textbf{State persistence.}
$I(W_n)$ is the set of particle IDs whose spheres intersect $W_n$; $D_n$ is the active set selected at control sample $n$ for the next interval, with $D_0=I(W_0)$. A particle outside the window remains dynamic until it is stable and outside the robot protection region. The update is
\begin{equation}
\label{eq:active-particle-set}
\begin{aligned}
D_n={}&I(W_n)\\
&{}\cup\{i\in D_{n-1}:\neg\operatorname{stable}_i^n\lor\operatorname{protected}_i^n\}.
\end{aligned}
\end{equation}
At control sample $n$, $\operatorname{stable}_i^n$ is true if speed and anchor-displacement bounds hold at all samples in the dwell interval; $\operatorname{protected}_i^n$ is true inside the robot margin or with a current robot contact. On exit, the runtime archives the particle state under its immutable global identifier. On reentry, it restores that record into a free solver slot. A bidirectional identifier to slot map enforces one active slot per global particle, so each dynamic particle is integrated once per physical substep. Archived regions remain frozen until reentry.

\section{Experiments}
We first assess the generated scenes through navigation planning and reinforcement learning-based policy training, followed by experiments on the efficiency and outcome consistency of task-adaptive simulation. Runtime denotes the elapsed execution time, whereas simulated time denotes the duration represented by the simulation. Hardware, software, and timing details are provided in Supplementary Sec.~G.

\noindent\textbf{Implementation details.}
We use GPT-6 Astra via ChatGPT for scene planning, rendered-view review, and task-conditioned physics configuration.

\subsection{Scene diversity and representation reuse}
RoboField generates ten environment categories with five visual configurations from L0 to L4 (Fig.~\ref{fig:scene-matrix}). Each column preserves a base scene and camera while geometry, materials, and decoration become progressively richer.

The shared scene identity allows each environment to be compiled for route planning and for contact, granular, fluid, or deformable interaction through the specialized backends in Fig.~\ref{fig:physics}.

\providecommand{\NavFigMain}{\textbf{Navigation planning across robots and task conditions.} (a) Planned UAV paths in a generated ruins scene, with an inset showing obstacle occupancy. (b) Planned quadruped paths in a generated forest scene, with an inset showing feasible ground-support regions. (c) Humanoid planning success under straight-corridor, goal-heading-change, and obstacle-detour conditions. (d) Worm planning success under short-visible, moderate-detour, and longer-or-narrow conditions. Point labels indicate successful/valid attempts.}

\providecommand{\NavTabMain}{\textbf{Navigation planning across five robot types.} Three methods share ten endpoint pairs per robot--scene combination (270 tasks; 810 attempts). Success is measured at the planning level. Median and 95th-percentile (P95) planning times include all attempts, including failures and timeouts, and any optimization after the first valid plan.}

\begin{figure}[t]\centering
\includegraphics[width=\columnwidth]{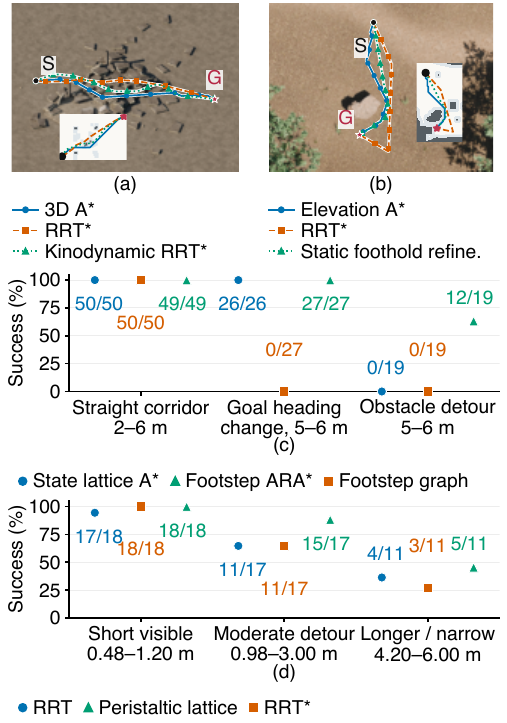}
\caption{\NavFigMain}\label{fig:navigation}
\end{figure}

\providecommand{\NavFigTime}{\textbf{Fraction of planning tasks solved over time.} Each task is counted at its first independently validated complete path. All valid attempts, including failures and timeouts, remain in the denominator. The plot includes the 810 attempts in main Table~\ref{tab:navigation} and excludes the additional humanoid and worm diagnostic suites in main Fig.~\ref{fig:navigation}(c,d). Times are measured from the start of planning and include independent verification. The horizontal axis uses a linear scale from 0 to 0.01~s and a logarithmic scale thereafter. The planning protocol is described in Sec.~\ref{sec:supp-nav}. Panels show (a) quadrupeds, (b) humanoids, (c) worms, (d) UAVs, and (e) underwater robots.}

\providecommand{\NavFigDistribution}{\textbf{Task geometry and planning outcomes under the original protocol.} (a) Boxes show median and quartiles, with whiskers at 1.5 IQR. All raw distances are overlaid. Each of the 270 endpoint tasks appears once. S/D/N denote Short, Detour, and Narrow or long, with counts printed below. Ground distances use XY. Aerial and underwater distances use XYZ. All distances are in metres. The original categories and documented exceptions for short distances are retained. (b) All 810 attempts are split into success, ordinary failure and timeout within each robot, method and original category. Numbers at the bar ends are denominators. Excluded or inapplicable combinations appear in Table~\ref{tab:snav-protocol} and are not added as failures. Geometric categories are not calibrated difficulty levels. These descriptive associations do not isolate distance as a cause.}

\providecommand{\NavTabProtocol}{\textbf{Task protocol and coverage.} EVAL denotes evaluated combinations. OUT denotes combinations excluded from this retained cohort. N/A denotes combinations outside the declared robot domain. $D/E$ count declared/qualified endpoint tasks. $A$ counts valid method attempts (three per endpoint). S/D/N are counts of the original geometric categories. Each scene has one frozen highest detail L4 instance. All evaluated rows use a 60 s planning runtime budget, 150 s hard runtime limit, one numerical thread, and the fixed seed. Ground admission retains complete motion envelopes and native support, slope, relief, phase or curvature rules. Aerial and underwater admission uses 3D occupancy and the corresponding motion constraints. OUT does not denote zero success. The CSV protocol table contains full robot envelopes, ROI, exclusions, scene versions, and endpoint IDs.}

\providecommand{\NavTabDetailed}{\textbf{Detailed planning performance and paired path quality.} Part A covers all valid attempts. It reports success/attempts, ordinary failures, timeouts, and complete planning runtime median/P95. Part B includes only identical endpoints where all three methods succeed. Results are pooled separately within each robot and reported as median [Q1,Q3]. Final polyline length is the retained geometric path length, not executed distance. Relative length is computed per task against the shortest observed final valid polyline among the compared methods, then summarized. It does not express theoretical optimality. Time to first accepted plan and final optimized path quality refer to different events. The validation margin bound uses frozen occupancy/support maps, erosion by the full envelope, and padding for applicable curved motion. It is not exact mesh distance or physical contact. The five common humanoid tasks support only limited conditional conclusions. All paired IDs and source memberships for each cell are exported.}

\providecommand{\NavTabDiagnostics}{\textbf{Humanoid and worm diagnostic experiments.} Rows are separate cohorts or explicitly identified reference views. They must not be summed as new independent executions. $D/E/A/S$ denote declared tasks, qualified tasks, valid attempts and successes per method. Fail denotes ordinary failure. TO denotes timeout. QG/RG denote qualification/resource gaps. Interface gaps are zero in the displayed cohorts and remain an explicit CSV field. All diagnostics use a 60 s runtime limit. Worm 4x changes native iteration/expansion limits, not the runtime limit or tracking cap. The original main rows provide context only. Humanoid nested targets reuse ten corridors. The original turning/detour aggregate overlaps its predeclared subsets for heading changes and detours. Short corridors remain reported here. Detours retain the original goal heading constraint unless explicitly marked as using position only. The worm revised view of 60 tasks references 153 unchanged original method records plus 27 runs that repair headings. These are not 180 new runs. The revised operational cohort reuses 62 identical results from the initial operational cohort and adds 232 results produced directly under the revised protocol. Qualified conditions, failures and gaps remain intact. All evidence here concerns planning or kinematic proxies, not full body execution.}

\providecommand{\NavFigMainDetails}{(a,b) Frozen ruins/UAV p03 and forest/quadruped p04 scenes show saved planned paths, shared task inputs for each method, start (S), goal (G), and canonical collision/support map insets. Camera overlays project recorded XYZ paths without depth testing. Ground XY paths are draped on the frozen terrain with a 0.055 m display offset. In (a)'s inset, routes are XY projections over the single slice at the starting altitude ($z=2.1$ m). Routes may leave that altitude. Neither a slice nor an overlay without depth testing certifies collision freedom along the full route. Insets show conservative planning geometry, not extra scene obstacles. No execution trajectory or unverified robot pose is shown. (c) Humanoid conditions include nested 2--6 m targets on ten straight corridors. They also include the predeclared subsets for goal heading changes and obstacle detours from the original 5--6 m batch. ARA* has one resource gap in straight corridors. Heading change has 30 declarations/27 qualified tasks per method, with 1 additional state lattice resource gap. Detour has 20 declarations/19 qualified tasks per method. The detour subset retains its original goal heading constraint. It does not use the later protocol that constrains position only. The 0.80--1.48 m short corridor results remain in the supplementary diagnostic table. (d) Worm task development strata contain 60 declared endpoints, 46 qualified endpoints, and 14 gaps in finite candidate search. Points label successes/valid attempts. Gaps are not algorithm failures. Conditions change across batches. This is neither a causal intervention on distance alone nor a test of unseen layouts. Diagnostics do not replace any of the 810 main cohort results.}

\subsection{Navigation planning and policy learning}
\label{sec:navigation-v2}

\noindent\textbf{Navigation planning.}
We evaluate quadruped, humanoid, worm, aerial, and underwater robots using planning representations that preserve scene geometry and incorporate robot-specific motion envelopes, support rules, and occupancy constraints. Three planning methods are selected for each robot type, with methods including A*~\cite{hart1968astar}, Anytime Repairing A* (ARA*)~\cite{likhachev2003ara}, and rapidly exploring random tree (RRT) variants, including asymptotically optimal RRT*~\cite{karaman2011rrtstar}. Ten fixed L4 scenes yield 27 robot--scene combinations, each with ten start--goal pairs shared by the three methods, giving 270 tasks and 810 attempts. Table~\ref{tab:navigation} reports planning success rates and runtime.

\begin{table}[t]
\caption{\NavTabMain}\label{tab:navigation}
\centering\small
\setlength{\tabcolsep}{3pt}
\begin{tabular*}{\columnwidth}{@{\extracolsep{\fill}}lrrrr@{}}
\toprule
Method & Success & Timeouts & \multicolumn{2}{c}{Time (s)} \\
 & $n/N$ (\%) & & Median & P95 \\
\midrule
\multicolumn{5}{@{}l}{\textbf{Quadruped}} \\
Elevation A* & 59/60 (98.3) & 0 & 0.10 & 1.15 \\
RRT* & 55/60 (91.7) & 0 & 6.72 & 26.74 \\
\shortstack[l]{Static foothold\\refine.} & 55/60 (91.7) & 0 & 0.17 & 1.44 \\
\midrule
\multicolumn{5}{@{}l}{\textbf{Humanoid}} \\
State lattice A* & 5/50 (10.0) & 40 & 60.00 & 60.00 \\
Footstep graph & 6/50 (12.0) & 39 & 60.00 & 60.00 \\
Footstep ARA* & 21/50 (42.0) & 24 & 60.00 & 60.01 \\
\midrule
\multicolumn{5}{@{}l}{\textbf{Worm}} \\
RRT & 16/60 (26.7) & 0 & 0.36 & 0.91 \\
RRT* & 18/60 (30.0) & 0 & 0.41 & 1.59 \\
Peristaltic lattice & 31/60 (51.7) & 0 & 0.58 & 17.08 \\
\midrule
\multicolumn{5}{@{}l}{\textbf{UAV}} \\
3D A* & 85/90 (94.4) & 2 & 1.43 & 31.48 \\
RRT* & 71/90 (78.9) & 0 & 12.36 & 48.83 \\
\shortstack[l]{Kinodynamic\\RRT*} & 85/90 (94.4) & 5 & 60.00 & 60.02 \\
\midrule
\multicolumn{5}{@{}l}{\textbf{Underwater}} \\
3D A* & 10/10 (100.0) & 0 & 1.74 & 6.98 \\
RRT* & 10/10 (100.0) & 0 & 21.98 & 45.32 \\
\shortstack[l]{Dynamics\\constrained (3DoF)} & 7/10 (70.0) & 0 & 5.57 & 13.48 \\
\bottomrule
\end{tabular*}

\end{table}

Figure~\ref{fig:navigation}(a,b) presents two representative cases: unmanned aerial vehicle (UAV) planning in ruins and quadruped planning in a forest. To investigate the lower planning success rates of humanoid and worm robots in the main evaluation, we conduct two additional test suites. Humanoid tests examine straight corridors, goal-heading changes, and obstacle detours, while worm tests cover short visible routes, moderate detours, and longer or narrower routes, as shown in Fig.~\ref{fig:navigation}(c,d), respectively. These tests characterize planning performance under different task conditions.

\noindent\textbf{Reinforcement learning.}
For UAV navigation, we train three high-level Soft Actor-Critic (SAC)~\cite{haarnoja2018sac} policies with different random seeds in a forest, using an existing low-level controller. On 20 held-out start--goal pairs in the same layout, the frozen policies achieve 20/20, 19/20, and 15/20 successes, compared with 9/20 for an analytic goal-pursuit baseline. For ruins adaptation, we continued training the forest-trained SAC policy's high-level velocity-command residual while keeping the low-level controller frozen. The adapted policy achieved 19/20 successes on held-out trials, compared with 12/20 for the analytic goal-pursuit baseline.

For worm motion control, proximal policy optimization (PPO)~\cite{schulman2017ppo} optimizes six central pattern generator (CPG) gait parameters in XPBD soil. After 5,120 environment steps, the policy passes all three straight-motion evaluations under small initial-pose perturbations of the same soil state, achieving a mean forward displacement of 30.074 mm over 4.70 s of simulated time. Maximum lateral deviation and heading error across the trials are 1.633 mm and $4.025^\circ$, respectively. The learned policy is reused in the subsequent moving-window study.

\begin{table*}[t]
\caption{\textbf{Runtime and agreement with numerical references.} (a) Excavation uses uniform fine configuration A as the reference. (b) Worm locomotion uses full-domain simulation as the reference. RMSE values are computed relative to the corresponding reference.}\label{tab:adaptive}
\centering
\small
\textbf{(a) Excavation: 32 s of simulated time}\par
\begingroup
\setlength{\tabcolsep}{4pt}
\begin{tabular}{@{}lrrrrrr@{}}
\toprule
Method & Diameter (mm) & Runtime (h) & \shortstack{Final bucket\\load (kg)} & \shortstack{Peak bucket\\load (kg)} & Load RMSE (kg) & Height RMSE (mm)\\
\midrule
A (reference) & 25 & 8.118 & 2.7317 & 3.1653 & \textemdash{} & \textemdash{}\\
B & 25/50 & 6.269 & 2.7100 & 3.1437 & 0.07679 & 26.10\\
C & 50 & 3.611 & 0.3469 & 0.3469 & 0.91750 & 79.56\\
\bottomrule
\end{tabular}
\endgroup
\par
\textbf{(b) Worm: 4.70 s in a shared world}\par
\begin{tabular}{lrrrrrr}
\toprule
Method & Runtime (s) & \shortstack{Mean active\\particles} & \shortstack{Peak active\\particles} & \shortstack{Forward\\Displacement (mm)} & \shortstack{3D COM Trajectory\\RMSE (mm)} & Shifts\\
\midrule
Full domain (reference) & 634.92 & 3,457,440 & 3,457,440 & 31.924 & \textemdash{} & 0\\
Moving window & 355.42 & 1,043,328 & 1,054,331 & 31.742 & 0.338 & 3\\
\bottomrule
\end{tabular}

\end{table*}

\begin{figure}[t]
\centering
\includegraphics[width=\columnwidth]{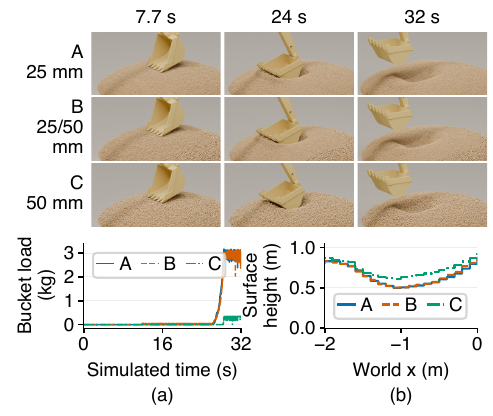}
\caption{\textbf{Excavation with fine, mixed, and coarse particle representations.} Rows show A (25 mm), B (25/50 mm with contact activation), and C (50 mm) at matched times. (a) Bucket load over simulated time. (b) Final terrain profile at $y=-0.15$ m.}\label{fig:excavation}
\end{figure}

\begin{figure}[t]
\centering
\includegraphics[width=\columnwidth]{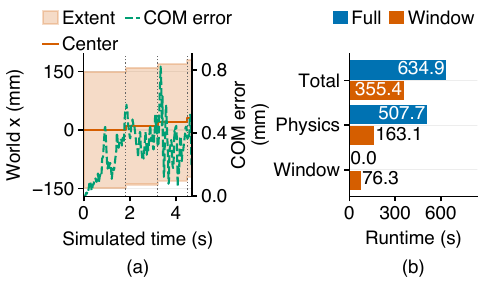}
\caption{\textbf{Moving-window simulation.} (a) Window state and instantaneous 3D COM position error relative to the full-domain reference. (b) Total runtime and selected component timings.}\label{fig:window}
\end{figure}

\subsection{Task-adaptive granular simulation}
We next evaluate task-adaptive simulation by examining its computational cost and agreement with numerical references during material interaction.

\noindent\textbf{Excavation.}
RoboField compiles the bucket interaction region as a granular material domain while retaining the surrounding scene as static collision and support geometry. Table~\ref{tab:adaptive} compares three configurations over the same 32 s bucket trajectory. A uses uniform 25 mm particles and provides the numerical reference. Mixed-resolution granular simulation in B concentrates 25 mm particles around the interaction region, uses 50 mm particles elsewhere, and activates the pile on contact. C uses uniform 50 mm particles. All configurations use XPBD granular updates.

The mixed-resolution run records 6.269 h of runtime versus 8.118 h for the fine numerical reference, a 22.77\% reduction. Its final bucket load is 2.7100 kg versus 2.7317 kg, a difference of 0.79\%. Load curve root-mean-square error (RMSE) is 0.07679 kg for the mixed-resolution run and 0.91750 kg for the coarse run (Fig.~\ref{fig:excavation}). The results characterize complete configurations with their recorded initial velocities and intervention policies.

\noindent\textbf{Moving active region.}
A second interaction case uses a worm robot to evaluate whether the computed granular region can move with the robot. We compare the window with full-domain simulation as the numerical reference in the same $0.60\times0.30$ m world. Both configurations share the same initial state of 3,457,440 particles and use the same frozen PPO policy (Sec.~\ref{sec:navigation-v2}) to simulate 4.70 s of worm locomotion. The window follows the robot through three translations and keeps 1,043,328 particles active on average, with a peak of 1,054,331 (Fig.~\ref{fig:window}).

Moving-window simulation reduces runtime from 634.92 s to 355.42 s (44.02\%), with a 3D center-of-mass trajectory RMSE of 0.338 mm relative to full-domain simulation. Final forward displacement is 31.742 mm for the moving window and 31.924 mm for the full domain.

\begin{table}[tb]
\caption{\textbf{Outcome agreement and cost of manipulation replay.} Agreement is measured against real outcomes. Runtime is summed over trajectories. Here, $n$ (S/F) gives the trajectory count and real success/failure counts.}\label{tab:real}
\centering\small
\begin{tabular}{@{}lrrrrr@{}}
\toprule
Layout & $n$ (S/F) & \multicolumn{2}{c}{\shortstack{Task Outcome\\Agreement}} & \multicolumn{2}{c}{Total Runtime (s)} \\
& & \multicolumn{1}{@{}c@{}}{Full} & \multicolumn{1}{@{}c@{}}{Selective} & \multicolumn{1}{@{}c@{}}{Full} & \multicolumn{1}{@{}c@{}}{Selective} \\
\midrule
1 & 20 (17/3) & 20/20 & 20/20 & 9761.84 & 8402.26 \\
2 & 22 (18/4) & 21/22 & 21/22 & 6408.42 & 5593.40 \\
3 & 22 (19/3) & 22/22 & 22/22 & 8424.68 & 6920.05 \\
\textbf{Overall} & \textbf{64 (54/10)} & \textbf{63/64} & \textbf{63/64} & \textbf{24594.94} & \textbf{20915.71} \\
\bottomrule
\end{tabular}

\end{table}

\begin{figure*}[t]
\centering
\includegraphics[width=0.96\textwidth]{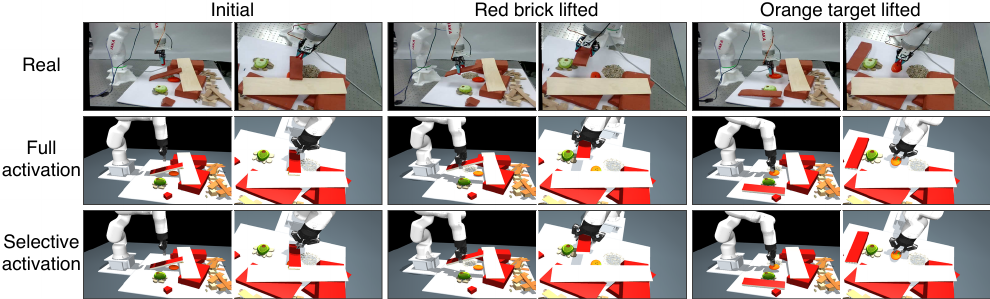}
\caption{\textbf{Paired replay of obstacle clearance and target retrieval.} Rows compare the real sequence, full activation, and selective activation. Columns show the initial state, obstacle lifting, and target retrieval, with two views at each stage.}\label{fig:real}
\end{figure*}

\subsection{Real-robot manipulation replay}
We evaluate object-level mixed-fidelity modeling by comparing dynamic simulation of all movable scene objects with task-conditioned dynamic--static assignment.

We use 64 recorded JAKA Mini2 trajectories across three tabletop layouts, including 54 successful and ten failed sequences. Tasks progress from direct retrieval to obstacle clearance before retrieval as target position, occlusion, and support relations vary (Fig.~\ref{fig:real}).

\noindent\textbf{Paired replay protocol.}
Measured joint feedback and gripper commands drive both modes. Full activation simulates every movable object. Selective activation simulates the task relevant objects and movable supports compiled before replay, while other geometry remains static. Both modes share the scene, motion, physical parameters, and stopping criterion. Task success requires lifting the target object and, for obstacle-clearance tasks, also removing the designated obstacle.

Both modes achieve task outcome agreement with real-world observations on 63/64 trajectories (98.44\%) and share the same mismatch (Table~\ref{tab:real}). Selective activation reduces total replay runtime from 24,594.94 s to 20,915.71 s, a 15.0\% saving. The task compiled scene therefore preserves outcome agreement while reducing replay computation.

\raggedbottom
\section{Conclusion}
We presented RoboField, connecting unstructured scene generation with task-adaptive simulation through a shared scene representation. Multimodal-model-guided scene planning, asset retrieval and on-demand generation, and procedural scene synthesis produce diverse environments with controllable visual detail for navigation planning and reinforcement learning. Mixed-fidelity modeling, contact activation, and moving windows concentrate computation on task-relevant interactions. Excavation, locomotion, and real-robot replay experiments demonstrate reduced runtime with close agreement to numerical references or observed task outcomes.

\section*{LLM Usage Statement}
OpenAI ChatGPT supported scene planning, rendered-view review, and physics configuration in RoboField. It helped develop code, design and run experiments, analyze results, and prepare data and figures. It also assisted with drafting, restructuring, editing, and translating the manuscript, reviewing literature and citations, and refining ideas, hypotheses, claims, and conclusions. The Asset Designer uses DeepSeek, the Visual Designer uses Doubao Seedream, and the 3D Modeler uses Hunyuan3D for on-demand asset generation. The authors reviewed all AI-assisted work and take full responsibility for all scientific decisions, results, and submitted content.

\par
{\small
\setlength{\bibsep}{0.5pt plus 1pt minus 0.5pt}

}
\clearpage
\setcounter{section}{0}
\setcounter{figure}{0}
\setcounter{table}{0}
\renewcommand{\thesection}{\Alph{section}}
\renewcommand{\thefigure}{S\arabic{figure}}
\renewcommand{\thetable}{S\arabic{table}}
\renewcommand{\theHsection}{supp.\arabic{section}}
\renewcommand{\theHfigure}{supp.\arabic{figure}}
\renewcommand{\theHtable}{supp.\arabic{table}}
\setcounter{topnumber}{3}
\setcounter{dbltopnumber}{4}
\renewcommand{\topfraction}{.93}
\renewcommand{\dbltopfraction}{.93}
\renewcommand{\textfraction}{.07}
\renewcommand{\floatpagefraction}{.68}
\renewcommand{\dblfloatpagefraction}{.68}
\providecommand*{\RoboFieldRevision}[1]{\textcolor{black}{#1}}
\makeatletter
\newcommand{\compactparagraph}{%
  \@startsection{paragraph}{4}{\z@}{.75ex \@plus .25ex \@minus .1ex}{-1em}%
    {\normalfont\normalsize\bfseries}}
\makeatother
\twocolumn[{%
  \begin{center}
    \pdfbookmark[0]{Supplementary Material}{supplementary}%
    {\Large\bfseries \RoboFieldTitle\\Supplementary Material\par}
    \vspace{2pt}
    {\large
      \begin{tabular}{c}
        \RoboFieldAuthors
      \end{tabular}\par}
    \vspace{0pt}
  \end{center}
}]
\raggedbottom

\section{Scene generation and fidelity}
\label{sec:supp-scene}

RoboField is an interactive simulation framework for robotics in unstructured environments. It integrates scene generation, physical simulation, and robot interaction through a shared scene representation. This section describes how environment requests are transformed into renderable scenes and a shared scene record $\mathcal S$, which captures the geometry, entities, and spatial relations needed by downstream tasks. Section~\ref{sec:supp-physics} then describes how this representation is compiled into task-specific physical models to enable interactive simulation.

\subsection{Inputs and scene planning}
\label{sec:supp-scene-inputs}
The scene request specifies the required landforms, objects, and spatial relations. Optional reference images guide appearance, and the spatial extent defines the construction region in metres. The requested visual level selects a modeling and rendering recipe. A fixed random seed controls stochastic placement and procedural variation.

The Scene Planner turns the request into SceneSpec, a structured description of scene entities, asset requirements, and spatial relations. Asset requirements guide the selection or construction of candidate geometry. Relations such as support, adjacency, and clearance guide the layout.

For the river scene, the user requests a continuous, open channel with vegetated banks. The Scene Planner encodes these requirements and the spatial relations among banks, water, and vegetation in SceneSpec to guide placement of the corresponding assets. The Scene Reviewer checks for crown encroachment and assesses channel visibility. If the layout fails the clearance check, affected vegetation is moved outside the checked corridor while preserving its visual detail at each tier. The revised layout is reviewed again.

\subsection{Asset Preparation and Quality Assurance}
SceneSpec's asset requirements guide catalog retrieval for objects with suitable semantics, licensed sources, and compatible dimensions and geometry. Geometric recipes can also supply terrain, repetitive structures, and placement patterns through procedural construction.

When retrieval cannot supply a required standalone asset, the Asset Designer turns the unmet request into an asset specification. The Visual Designer uses that specification to generate a reference image, which the 3D Modeler uses to produce a candidate mesh.

Before assembly, asset quality assurance checks semantic fit, metre scale, axes and local origin, geometric integrity, material dependencies, and provenance. Candidates may be repaired, replaced, referred for manual review, or rejected. Assets that pass the checks provide the geometry and placement attributes used with the spatial relations in SceneSpec during environment construction.

Infinigen~\cite{suppbib-infinigen} supplies procedural vegetation, and Poly Haven provides CC0 rocks and textures~\cite{suppbib-supp_polyhaven}. Ruin fragments include Japhet Mark's \textit{Concrete \RoboFieldRevision{Debris}} scan from BlenderKit under its Royalty Free license~\cite{suppbib-supp_blendkit}. The mine rail assembly is generated in-house.

\subsection{Environment-specific construction}
The construction program uses the checked assets and SceneSpec relations to build terrain and determine mesh variants, scales, transforms, and placements. Environment-specific construction preserves characteristic geometry, topology, and spatial relations through geometric constraints and placement checks. Table~\ref{tab:supp-recipes} summarizes the construction principles for the ten environment categories.

\paragraph{River construction example.}
The open-channel requirement in SceneSpec guides crown-aware vegetation placement. The layout program evaluates the bank envelope at 25 longitudinal sample positions within each L4 crown's span along the 112\,m reach. For vegetation that needs to move, the required crown-to-bank buffer is 2.5\,m for trees and 0.75\,m for shrubs. The corresponding minimum root-to-centerline distances are 13.5 and 7.0\,m. The resulting horizontal positions guide vegetation placement during scene assembly.

\begin{table*}[!tb]
\centering\small
\caption{Construction principles and geometric constraints for the ten environment categories.}\label{tab:supp-recipes}
\begin{tabular}{@{}p{2.1cm}p{5.5cm}p{8.5cm}@{}}\toprule
Environment & Base geometry and assets & Layout and checks\\\midrule
Forest & Continuous ground, connected shrub and tree meshes, understory & Maintain ground continuity and vegetation distribution, with consistent placement and support relations.\\
Mountain & Co-registered swisstopo swissALTI3D elevation (2024) and SWISSIMAGE orthoimagery (2023) near the Rh\^one Glacier~\cite{suppbib-supp_swisstopo}, with local rock meshes & Preserve terrain morphology, including ridges, cliffs, and road connectivity.\\
Desert & Procedural dune relief and source sand materials & Control dune morphology, ridge distribution, and slope transitions.\\
Glacier & Crevasse and ice/snow terrain meshes, source rock debris & Preserve crevasse geometry, ice-surface morphology, and surrounding terrain structure.\\
Wetland & Native irregular saturated banks, vegetation and shallow water appearance & Preserve bank and pool topology and maintain consistent vegetation placement.\\
Cave & Joint-bounded cave shell, floor pool, and breakdown-rock templates & Maintain passage connectivity, cavity geometry, and rock placement.\\
Mine & Mine tunnel, procedural rails and sleepers, ballast and walls & Maintain tunnel continuity, rail alignment, and structural relations.\\
Ruins & Native hall foundations, broken slabs, BlenderKit concrete fragments and rebar & Preserve structural geometry, rubble distribution, and support relations.\\
River & Continuous channel, procedural banks and source vegetation & Maintain channel continuity, bank geometry, and vegetation clearance.\\
Underwater & Reef and seabed geometry, retrieved underwater meshes and materials & Preserve seabed morphology, reef geometry, and open-water spaces.\\\bottomrule
\end{tabular}
\end{table*}

\subsection{Assembly and iterative review}
The assembly program instantiates the checked assets at their prescribed placements and renders fixed review views. The Scene Reviewer compares those views with the original scene requirements and proposes revisions when the layout or appearance falls short. Separate geometric checks test support, intersections, clearance, or connectivity as required by the scene recipe. Findings can lead to revisions of assets, placements, or geometry, followed by another assembly and review pass. Persistent failures are referred for manual revision or rejected.

\paragraph{River layout review.}
Review views use the same camera at $1200\times900$ pixels across visual tiers. Blender Cycles renders L0--L4 at 32, 64, 128, 256, and 768 samples per pixel, respectively. The corrected layout uses the same horizontal placements across tiers, with root origins 0.03\,m below the local terrain surface. The bank geometry is preserved. The reviewed assembly provides a renderable environment and the realized entity bindings used in the shared scene record.

\subsection{Shared scene representation}
Rendering, navigation planning, and physics compilation need a consistent way to identify entities in the realized assembly. SceneSpec describes the planned entities, asset requirements, and relations. The shared scene record $\mathcal S$ links those identities and relations to the concrete source meshes, transforms, and material and physical fields of the realized scene. A native scene can supply its existing mesh and pose bindings directly. The record uses a right-handed frame with $+Z$ up and $+X$ forward, lengths in metres, masses in kilograms, and normalized $wxyz$ pose quaternions. Validation resolves identifiers and references and checks dimensions, units, and pose consistency. Task-specific checks evaluate conditions such as adequate support.

\begin{figure*}[!tb]
\centering\includegraphics[width=\textwidth]{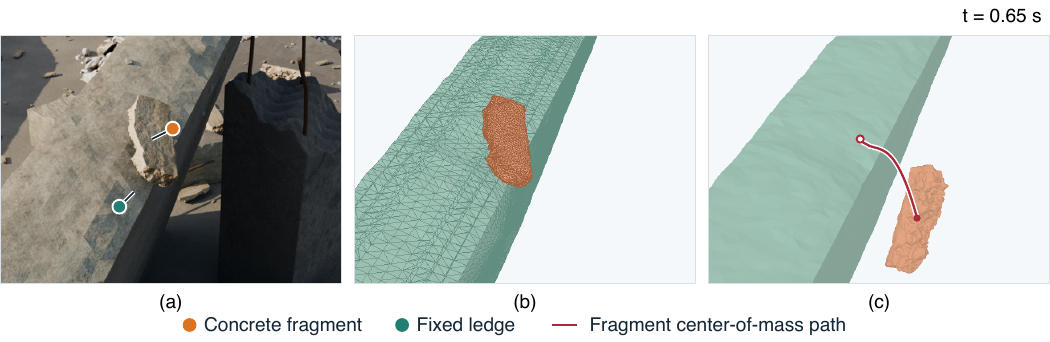}
\caption{Shared entity bindings for a concrete fragment and its supporting ledge in the ruins (main Secs.~3.1--3.2). (a) The initial scene render. (b) The source triangles used for signed-distance collision at the same initial pose. (c) The post-release rigid-body state at simulation time 0.65\,s, with the fragment's center-of-mass path from release. All three panels use the same camera and crop.}\label{fig:supp-representation}
\end{figure*}

Figure~\ref{fig:supp-representation} follows a scanned concrete fragment and its supporting ledge in the ruins. Their entity IDs in $\mathcal S$ link the rendered instances to the source triangles used for signed-distance collision. For the release task, the backend assigns the fragment to a movable rigid body and the ledge to static support. The fragment keeps its source mesh and initial world transform on the ledge. The compiled rigid-body origin is $(5.276,-2.362,1.449)$\,m, and its mesh-to-body transform carries the native orientation. Its mass is 2.020\,kg, with inertia computed from the collision representation. After release, the fragment's binding maps the simulated rigid-body pose back to its visual instance (Sec.~\ref{sec:supp-physics-compile}).

\subsection{Visual configurations}
The visual level selected in Sec.~\ref{sec:supp-scene-inputs} controls the appearance and visual geometry during construction and rendering. Table~\ref{tab:fidelity} lists the default L0--L4 recipes. The mesh ratio targets the retained face fraction of eligible visual meshes relative to their source meshes. Topology, open passages, and essential components can take precedence over the target. The texture cap limits the longest texture dimension when downsampling applies. The presets also control secondary placement density, shading contrast, roughness variation, and fine normal detail.

Main Fig.~\ref{fig:scene-matrix} qualitatively illustrates changes in geometry, materials, and decoration across L0--L4. Each column shows variants of the same base scene from a fixed viewpoint.

\begin{table*}[!tb]
\caption{Shared visual-fidelity recipes. Mesh fractions and texture caps are default targets for eligible visual assets.}\label{tab:fidelity}
\centering\small
\begin{tabular}{@{}lp{9.3cm}rr@{}}\toprule
Level & Visual recipe meaning & Mesh ratio & Texture cap (px)\\\midrule
L0 & Strong simplification, clay major surfaces, retained foliage silhouette & 0.008 & 64\\
L1 & Coarse colored geometry, subdued albedo, uniform roughness, sparse decoration & 0.02 & 64\\
L2 & Textured main structure, medium geometry, limited decoration, no fine normal map detail & 0.10 & 256\\
L3 & Coherent physically based materials and substantial secondary detail & 0.40 & 1024\\
L4 & Native geometry, source materials and full source decoration & 1.00 & Uncapped\\\bottomrule
\end{tabular}
\end{table*}

Section~\ref{sec:supp-physics} uses the shared scene record to derive task-specific physical representations and discretizations.

\section{Physics integration}
\label{sec:supp-physics}

Section~\ref{sec:supp-scene} provides a renderable environment and the shared scene record $\mathcal S$. We combine $\mathcal S$ with the robot, task, and initial state to configure the interaction. Sec.~\ref{sec:supp-physics-compile} describes \RoboFieldRevision{the compilation of} dynamic states and static boundaries\RoboFieldRevision{, as well as the mapping of} solver outputs back to scene entities. Sec.~\ref{sec:supp-physics-cases} details the numerical models and execution modes.

\subsection{Compilation and state transfer}
\label{sec:supp-physics-compile}
The Physics Configurator uses the shared scene record, the robot geometry, the task, and the initial state to propose task-relevant entities and representation categories. Recorded or prescribed robot motion also informs selection when available. Deterministic case code resolves the proposed IDs, follows support relations, and checks the region swept by the robot or tool to include geometry needed for contact and support. The code checks geometry and material fields and assembles the solver input. Case templates select the solver and specify the numerical settings, including constitutive parameters, discretization, initial conditions, loads, constraints, and update schedules. Where applicable, these settings use extracted native geometry or measured and simulated load records. Authored material values characterize the corresponding numerical case.

Backend adapters bind the selected entity IDs and coordinate frames to dynamic solver states and static collision and support boundaries. Rigid models use source triangles, signed-distance collision, or convex proxies as appropriate. Volume samples initialize particle domains, while source surfaces and skeletal geometry bound soil finite elements and beam cross-sections. The entity bindings also carry the solver output back to the scene. Rigid poses update source instances during replay, nodal displacements move attached surface geometry, and particle positions reconstruct material surfaces. These states also support task metrics such as trajectory error, load curves, and contact response.

\begin{table*}[!tb]
\centering\small
\caption{Solver backends, execution, and outputs for the physical examples in main Fig.~\ref{fig:physics}.}\label{tab:backends}
\begin{tabular}{@{}p{2.5cm}>{\raggedright\arraybackslash}p{5.35cm}p{8.25cm}@{}}\toprule
Scene & Solver backends and roles & Execution and output\\\midrule
Desert & Newton~\cite{suppbib-newton} (sand MPM)\newline MuJoCo~\cite{suppbib-mujoco} (robot dynamics) & Partitioned articulated/material solve with velocity and wrench exchange. Particle, surface and robot states.\\
Forest & SOFA~\cite{suppbib-faure2012sofa} (static branch BeamFEM)\newline MuJoCo (robot stance) & Quasistatic branch response, prescribed tool load and separate robot stance. Deformed beam and surface geometry.\\
Wetland & MuJoCo (\RoboFieldRevision{robot stance on rigid ground})\newline Kratos~\cite{suppbib-dadvand2010kratos} (porous soil, $u$--$p$ FEM) & Foot loads from the MuJoCo stance drive the Kratos solve for soil displacement and pore pressure. Surface displacements determine the compatible robot pose.\\
Mine & Chrono~\cite{suppbib-tasora2016chrono} (NSC rail contact) & Rigid contact of an unfastened rail under an applied lateral force history. Rail and contact states.\\
Cave and river & SPlisHSPlasH~\cite{suppbib-splishsplash,suppbib-bender2017dfsph} (DFSPH fluid and PBD rigid boundaries) & Local particle--rigid interaction. Particle positions reconstruct water surfaces, and rigid poses update the scene objects.\\
Underwater & Gazebo~\cite{suppbib-gazebo_hydrodynamics} (buoyancy and drag)\newline DART~\cite{suppbib-lee2018dart} (dynamics with added mass) & Passive ROV drift in still water and uniform flow. Vehicle position, orientation and velocity.\\
Glacier, mountain & MuJoCo (rigid contact) & Friction-dependent footing on ice and a separately simulated gravity-driven rockfall. Motion and contact states.\\
Ruins & MuJoCo (rigid bodies, SDF contact) & Release of an initial pose constraint followed by gravity and rigid contact. Fragment and contact states.\\\bottomrule
\end{tabular}
\end{table*}

\subsection{Numerical models and cases}
\label{sec:supp-physics-cases}
\begin{figure*}[!t]
\centering\includegraphics[width=\textwidth]{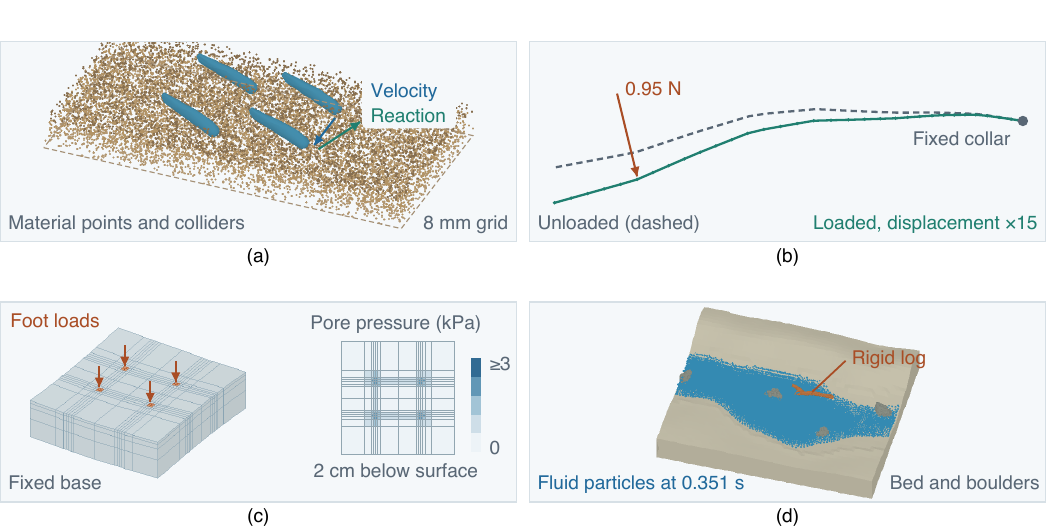}
\caption{Solver domains for four physical interaction examples in main Fig.~\ref{fig:physics}. (a) Desert sand with two-way MPM coupling. Sampled material points and foot/calf collision geometry at 6\,s. The schematic arrows indicate the two-way exchange of velocity and reactions. (b) Forest branch under a prescribed load. Beam nodes before and after a prescribed 0.95\,N load. Displacements are magnified $15\times$. (c) Wetland footing with one-way finite-element loading. The mesh, four loaded patches and excess pore pressure 2\,cm below the surface at 0.5\,s. Color indicates the change in pore-water pressure relative to the initial equilibrium state. Darker shades indicate a larger pressure increase. (d) River reach with local fluid--rigid interaction. Fluid particles, the rigid log and the fixed boundaries at 0.351\,s.}\label{fig:supp-domains}
\end{figure*}

In Table~\ref{tab:backends}, MPM denotes the material point method, DFSPH divergence-free smoothed particle hydrodynamics, PBD position-based dynamics, FEM the finite element method, and NSC nonsmooth contact. BeamFEM uses beam finite elements. In $u$--$p$ FEM, $u$ is solid displacement and $p$ is pore pressure. The following cases show how task-relevant entities from the shared scene are compiled into numerical solver domains with distinct execution modes. Figure~\ref{fig:supp-domains} depicts four such domains, spanning two-way exchange, prescribed loading, and local fluid--rigid interaction.

\paragraph{Desert sand.}
A native dune patch provides the initial sand surface. The sand has 2,647,118 material points with 4\,mm spacing and an 8\,mm grid. The template uses density $1600\,\mathrm{kg/m^3}$, friction angle $30^\circ$, Young's modulus 10\,MPa and Poisson ratio 0.3. The base of the local patch supports the granular material\RoboFieldRevision{, while} the robot feet receive support through the material solver. Starting from a settled free-base pose, a torque-limited crawl moves the four foot/calf colliders. Every 1\,ms a MuJoCo predictor supplies collider poses and velocities to implicit MPM. Grid-contact and projection impulses are grouped by body, converted to forces and moments, and applied in a corrected articulated step after restoring the pre-prediction state. Material-point positions reconstruct the deformed sand surface, and articulated states determine the robot poses.

\paragraph{Forest branch.}
The original branch centerline and fitted cross-sections define a 33-node beam with a fixed 20\,mm anatomical collar. The unloaded shape serves as the reference. SOFA's static Newton--Raphson solver advances a sequence of equilibria under a prescribed normal probe load, reaching 0.95\,N in the image case. Beam-section translations and rotations map bark and leaves back to the native geometry. The beam tip displacement is 1.157\,mm under the prescribed load and authored elastic stiffness. Separately, MuJoCo computes the robot stance under the corresponding reaction wrench applied to the tool.

\paragraph{Wetland footing.}
MuJoCo simulates the Go2 stance on rigid ground. The four three-dimensional foot forces are averaged over 5--6\,s, after the stance has settled. Kratos applies these loads at four disjoint footprints on a $1\times1\times0.3$\,m source-surface soil patch. Each footprint covers four 2\,cm surface cells. Assembled surface loads preserve the resultants and application centers of the stance loads. Kratos uses 784 mixed-order 20-node hexahedra, with displacement at all nodes and pore pressure at eight corners. The incremental state starts at zero excess pore pressure about a prebalanced saturated reference. Authored parameters include Young's modulus 2\,MPa, Poisson ratio 0.30, porosity 0.45, and isotropic permeability $10^{-12}\,\mathrm{m^2}$. Bottom displacement and side-normal displacement are fixed. Excess pressure is fixed to zero on the exposed top outside the impermeable footprints\RoboFieldRevision{, while} other hydraulic boundaries have zero natural flux. Loads ramp to their means in 0.5\,s and are then held. Time steps are 0.25\,s up to 1\,s and 1.0\,s thereafter. Fields at 0.5 and 2.0\,s supply displacement and pressure. Inverse kinematics places the four feet at the displaced contact locations.

\paragraph{Mine rail.}
In Chrono NSC, a rigid rail assembled from its head, web, and foot interacts with the sleepers, ballast, and floor. To establish the unfastened configuration shown in main Fig.~\ref{fig:physics}, 50 fastening components, including fasteners and fishplates, are omitted from the physical model, while the rail geometry and inertial properties are retained. The nominal friction coefficient is 0.35, with a 0.5\,ms time step. A prescribed lateral force ramps up from 0.5 to 0.6\,s, remains at 300\,N until 0.7\,s, and is removed by 0.8\,s. The resulting rail motion and contact reactions characterize the response of the unfastened assembly.

\paragraph{Cave pool and river reach.}
In the cave, an initially stationary finite water pool is bounded by source floor geometry. A source rock fragment drops under gravity and interacts with the free surface through a dynamic PBD boundary. The refined local run used particle radius 1.5\,mm, spacing 3\,mm and a 0.125\,ms step, with water-like kinematic viscosity and a surface-energy model. Fluid particle positions reconstruct the water surface, and rigid transforms position the rock mesh. In the river, a 29.91\,kg native log is released in a bounded water reach contacting the bed, shore and four source boulders. The DFSPH/PBD mesh-boundary simulation uses 14,036 particles of 40\,mm radius and a 1\,ms step. The initial finite water volume evolves within the native boundaries. The same reconstruction maps fluid and rigid states to the visible water surface and log. Reaction estimates derived from body momentum include non-fluid contact contributions.

\compactparagraph{Passive underwater drift.}
The BlueROV2 starts at the same pose under zero thrust in both cases. Gazebo supplies buoyancy for water density $1000\,\mathrm{kg/m^3}$ and hydrodynamic drag, with source added-mass terms supplied to DART. The comparison uses still water and a uniform $+X$ current of 0.35\,m/s, with 1\,ms steps over 3\,s. The reef provides the source context and a clearance check. Vehicle poses and velocities quantify passive drift.

\paragraph{Glacier and mountain.}
The glacier compares two 2\,s repeats of the same initial Go2 pose on rigid native ice terrain using the same torque-limited stance controller. The friction coefficients are 0.80 and 0.03. Feet and base motion quantify sliding. In the mountain scene, a source rock is released above the native Swiss terrain. Local terrain triangles are extruded downward for collision support and surrounding source convex pieces form fixed obstacles. The rigid contact solver integrates the fall, impacts, and subsequent roll over 3\,s at a 0.5\,ms step. The resulting rigid poses drive the visual rock mesh.

\paragraph{Ruins release.}
A 2.020\,kg concrete fragment starts in its retained transform on a native ledge. Sixty-two other local objects are fixed. Source triangle signed distances provide contact. At 0.25\,s\RoboFieldRevision{,} the prescribed pose hold is released\RoboFieldRevision{, after which} gravity and contact determine the resulting motion. The solver computes the fragment pose, velocity, and contact response after release.

\section{Navigation details}
\label{sec:supp-nav}
\begin{table}[!t]
\caption{Coverage of the main planning cohort. Each evaluated combination uses one fixed L4 scene, ten eligible start--goal pairs shared by three methods, one numerical thread, and a fixed random seed. 60~s planning budget and 150~s end-to-end wall-clock limit. \(\bullet\) denotes an evaluated combination. \textemdash{} denotes ``Not evaluated'' and covers out-of-scope and inapplicable combinations, all omitted from success-rate denominators.}\label{tab:snav-protocol}
\centering\small
\setlength{\tabcolsep}{2.5pt}
\begin{tabular}{@{}lccccc@{}}\toprule
Scene & UAV & Quadruped & Worm & Humanoid & Underwater \\
\midrule
Forest & \(\bullet\) & \(\bullet\) & \(\bullet\) & \(\bullet\) & \textemdash{} \\
Mountain & \(\bullet\) & \(\bullet\) & \(\bullet\) & \(\bullet\) & \textemdash{} \\
Desert & \(\bullet\) & \(\bullet\) & \(\bullet\) & \(\bullet\) & \textemdash{} \\
Ruins & \(\bullet\) & \(\bullet\) & \(\bullet\) & \(\bullet\) & \textemdash{} \\
River & \(\bullet\) & \(\bullet\) & \(\bullet\) & \(\bullet\) & \textemdash{} \\
Wetland & \(\bullet\) & \(\bullet\) & \(\bullet\) & \textemdash{} & \textemdash{} \\
Cave & \(\bullet\) & \textemdash{} & \textemdash{} & \textemdash{} & \textemdash{} \\
Mine & \(\bullet\) & \textemdash{} & \textemdash{} & \textemdash{} & \textemdash{} \\
Glacier & \(\bullet\) & \textemdash{} & \textemdash{} & \textemdash{} & \textemdash{} \\
Underwater & \textemdash{} & \textemdash{} & \textemdash{} & \textemdash{} & \(\bullet\) \\
\bottomrule
\end{tabular}
\end{table}

Table~\ref{tab:snav-protocol} gives the coverage of the 810 planning attempts reported in main Table~\ref{tab:navigation}. For ground robots, the admission checks use their motion envelopes and applicable support, slope, relief, phase, and curvature constraints. For UAV and underwater robots, they use 3D occupancy and the corresponding motion constraints.

\paragraph{Geometry and admissibility.}
The 27 evaluated robot--scene combinations use 270 distinct start--goal tasks and three methods per task. Maps are derived from the frozen L4 source geometry. Ground maps store support height and clearance from overhead and blocking geometry, and aerial and underwater maps store 3D occupancy. Raster spacing is case specific. The planner tests the robot's full motion envelope against this geometry. Endpoint pairs are selected in connected admissible regions as visible short tasks, blocked line-of-sight detours, or narrow/long tasks. The independent verifier checks the start and goal tolerances, swept geometry support and collisions, and applicable heading, curvature, gait phase or dynamic limits along the entire returned path.

\paragraph{Planning runtime.}
Complete planning runtime covers the full planning stage after setup, including independent validation and any search that continues after a path is accepted.

Each planning attempt is subject to two time limits. The nominal 60~s planning budget starts after map preparation and covers path search, independent validation, and any optimization that continues after the first path is validated. Recorded planning times can slightly exceed this budget during deadline handling or when a native planner finishes its current iteration and the adapter verifies the returned path. Separately, a 150~s end-to-end wall-clock limit starts when the worker process is launched and covers initialization, map preparation, planning, and result writing. An outer watchdog terminates the worker if it exceeds this limit.

\begin{figure*}[!tb]
\centering
\includegraphics[width=\textwidth]{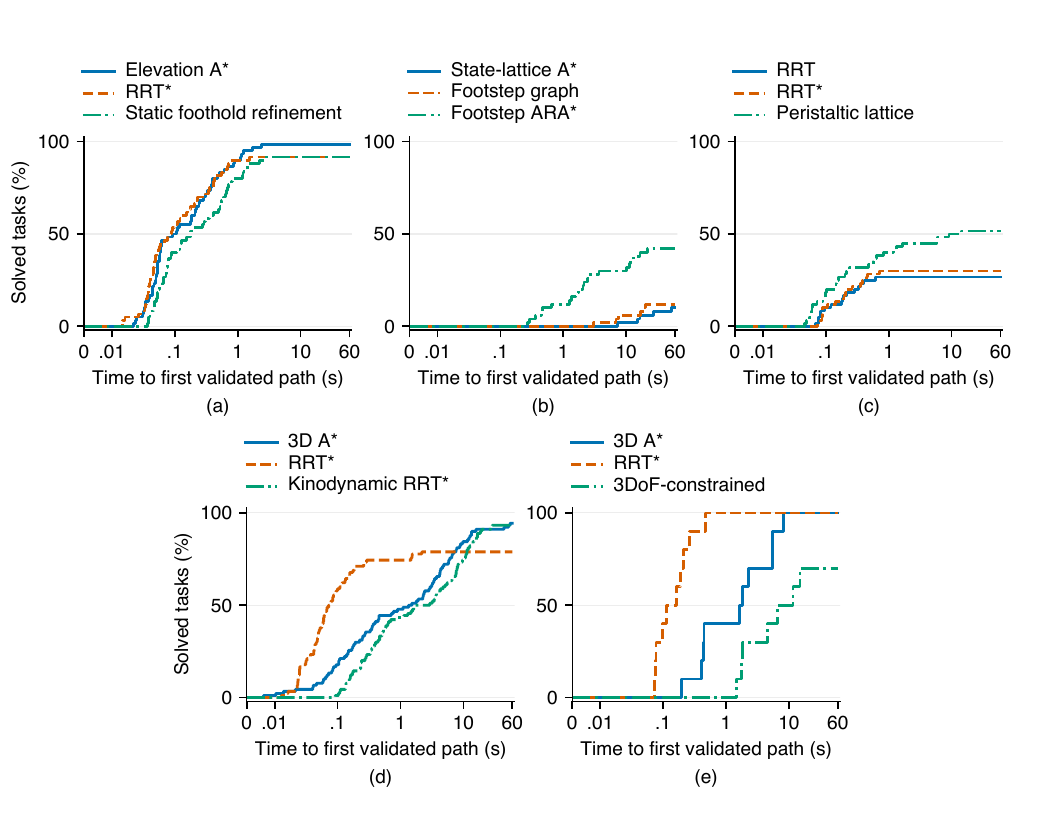}
\caption{\NavFigTime}\label{fig:snav-time}
\end{figure*}

Figure~\ref{fig:snav-time} measures from the planning start and records when the first complete path passes independent validation. Archival processing is timed separately. The total runtime distribution pools all 810 raw attempts, giving a median of 4.30\,s and P95 of 60.00\,s. Table~\ref{tab:snav-detailed} gives success, timeout, and complete runtime statistics per evaluated combination.

\paragraph{Main figure cases and diagnostic strata.}
Main Fig.~\ref{fig:navigation}(a,b) shows planned paths and map projections for two representative cases. The ruins UAV case in (a) uses a voxel grid with 0.50\,m spacing, and the forest quadruped case in (b) uses a ground grid with 0.10\,m spacing. Table~\ref{tab:snav-diagnostics} summarizes the six conditions in panels (c,d). The humanoid straight condition uses nested targets on ten corridors, and the obstacle detour retains its goal heading constraint. The worm categories describe the task geometry. They are evaluated separately from the main cohort.

\begin{table*}[!tb]
\caption{Planning outcomes and runtime by robot and scene. Each method cell gives successes, timeouts, and median / P95 elapsed planning time in seconds. Each method has ten attempts per evaluated combination. Ordinary failures are ten minus successes and timeouts. Timing includes all outcomes.}\label{tab:snav-detailed}
\centering\small
\begin{tabular}{@{}lccc@{}}\toprule
\textbf{Quadruped} & Elevation A* & RRT* & Foothold refinement \\
Forest & 10, 0, 0.05 / 0.08 & 10, 0, 5.47 / 9.69 & 10, 0, 0.06 / 0.11 \\
Mountain & 10, 0, 0.23 / 0.35 & 10, 0, 9.03 / 11.77 & 10, 0, 0.51 / 0.76 \\
Desert & 10, 0, 0.29 / 1.86 & 10, 0, 8.77 / 25.77 & 8, 0, 0.42 / 2.09 \\
Wetland & 10, 0, 0.05 / 0.06 & 7, 0, 2.42 / 2.95 & 9, 0, 0.06 / 0.07 \\
Ruins & 9, 0, 0.50 / 1.46 & 9, 0, 9.77 / 27.32 & 9, 0, 0.75 / 1.90 \\
River & 10, 0, 0.27 / 0.61 & 9, 0, 10.03 / 35.58 & 9, 0, 0.38 / 0.74 \\
\midrule
\textbf{Humanoid} & State lattice A* & Footstep graph & Footstep ARA* \\
Forest & 3, 6, 60.00 / 60.00 & 3, 6, 60.00 / 60.00 & 5, 4, 60.00 / 60.00 \\
Mountain & 0, 10, 60.00 / 60.00 & 0, 10, 60.00 / 60.00 & 3, 7, 60.01 / 60.02 \\
Desert & 0, 6, 60.00 / 60.00 & 0, 6, 60.00 / 60.00 & 6, 0, 60.00 / 60.01 \\
Ruins & 0, 10, 60.00 / 60.00 & 0, 10, 60.00 / 60.00 & 2, 8, 60.00 / 60.01 \\
River & 2, 8, 60.00 / 60.00 & 3, 7, 60.00 / 60.00 & 5, 5, 60.00 / 60.01 \\
\midrule
\textbf{Worm} & RRT & RRT* & Peristaltic lattice \\
Forest & 4, 0, 0.24 / 0.45 & 5, 0, 0.25 / 0.49 & 6, 0, 11.78 / 16.89 \\
Mountain & 0, 0, 0.88 / 0.92 & 0, 0, 1.46 / 1.61 & 5, 0, 11.13 / 17.65 \\
Desert & 3, 0, 0.75 / 0.78 & 3, 0, 1.07 / 1.12 & 6, 0, 0.74 / 16.96 \\
Wetland & 3, 0, 0.26 / 0.29 & 4, 0, 0.26 / 0.30 & 5, 0, 0.03 / 0.70 \\
Ruins & 3, 0, 0.47 / 0.62 & 3, 0, 0.54 / 0.80 & 3, 0, 16.61 / 17.11 \\
River & 3, 0, 0.33 / 0.36 & 3, 0, 0.37 / 0.39 & 6, 0, 0.05 / 0.18 \\
\midrule
\textbf{UAV} & 3D A* & RRT* & Kinodynamic RRT* \\
Forest & 10, 0, 0.49 / 3.66 & 4, 0, 0.64 / 9.84 & 10, 0, 60.00 / 60.42 \\
Mountain & 8, 2, 28.52 / 60.35 & 4, 0, 1.42 / 60.01 & 6, 4, 60.01 / 60.33 \\
Desert & 10, 0, 4.28 / 10.53 & 10, 0, 19.18 / 40.93 & 10, 0, 60.00 / 60.00 \\
Glacier & 10, 0, 3.56 / 13.96 & 10, 0, 28.82 / 42.81 & 10, 0, 60.00 / 60.00 \\
Wetland & 10, 0, 2.79 / 11.84 & 10, 0, 17.50 / 32.60 & 10, 0, 60.00 / 60.00 \\
Cave & 7, 0, 1.21 / 9.92 & 10, 0, 31.10 / 49.44 & 10, 0, 60.00 / 60.00 \\
Mine & 10, 0, 0.20 / 0.42 & 10, 0, 10.72 / 19.19 & 10, 0, 60.00 / 60.00 \\
Ruins & 10, 0, 0.10 / 0.62 & 10, 0, 9.52 / 15.89 & 10, 0, 60.00 / 60.00 \\
River & 10, 0, 3.26 / 7.57 & 3, 0, 0.68 / 16.75 & 9, 1, 60.00 / 60.00 \\
\midrule
\textbf{Underwater} & 3D A* & RRT* & Dynamics-constrained 3DoF \\
Underwater & 10, 0, 1.74 / 6.98 & 10, 0, 21.98 / 45.32 & 7, 0, 5.57 / 13.48 \\
\bottomrule
\end{tabular}
\end{table*}

\begin{table*}[!tb]
\caption{Planning diagnostics corresponding to main Fig.~\ref{fig:navigation}(c,d). Entries report successes out of valid attempts. Eligible endpoint counts vary by condition. Resource limits left one eligible pair without a valid attempt for Footstep ARA* in straight corridors and one for State lattice A* in goal-heading changes.}\label{tab:snav-diagnostics}
\centering\small
\begin{tabular}{@{}lccc@{}}\toprule
\textbf{Humanoid} & State lattice A* & Footstep graph & Footstep ARA* \\
Straight corridor, 2--6 m & 50/50 & 50/50 & 49/49 \\
Goal heading change, 5--6 m & 26/26 & 0/27 & 27/27 \\
Detour with goal heading, 5--6 m & 0/19 & 0/19 & 12/19 \\
\midrule
\textbf{Worm} & RRT & RRT* & Peristaltic lattice \\
Short visible & 17/18 & 18/18 & 18/18 \\
Moderate detour & 11/17 & 11/17 & 15/17 \\
Longer or narrow & 4/11 & 3/11 & 5/11 \\
\bottomrule
\end{tabular}
\end{table*}

\section{Robot learning and control}
\label{sec:supp-learning}
\begin{figure*}[!tb]
\centering\includegraphics[width=\textwidth]{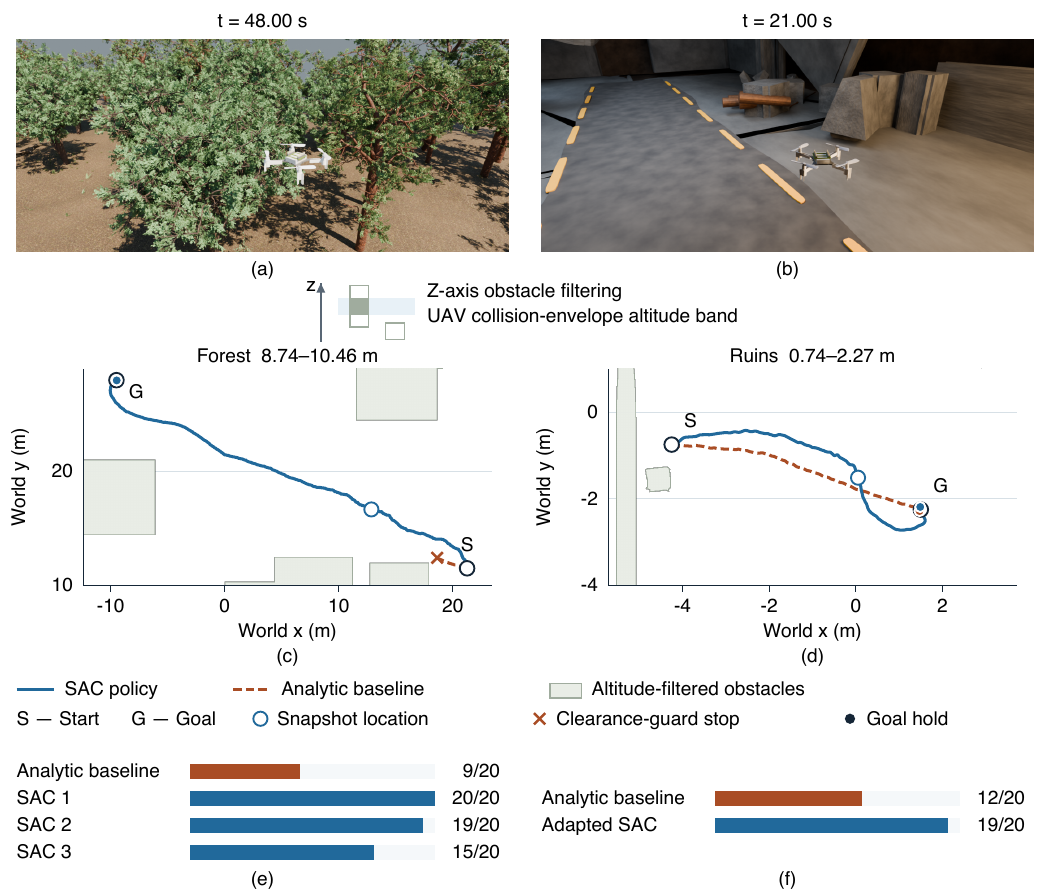}
\caption{UAV policy evaluation in forest and ruins (main Sec.~\ref{sec:navigation-v2}). (a) Logged Crazyflie pose in the L4 forest scene at 48.00\,s. (b) Logged Crazyflie pose in the original ruins visual scene at 21.00\,s. (c) Forest and (d) ruins XY projections of 3D trajectories executed in simulation by the analytic baseline and SAC policy from matched start and goal positions. For each representative matched flight pair, the indicated altitude band spans the UAV collision envelopes across both flights. The central schematic illustrates the common filtering rule. Obstacle footprints project the forest collision boxes intersecting this band and the portions of the ruins collision meshes within it. XY overlap can combine different heights and times. Crosses indicate a clearance-guard stop, and solid dots indicate completion of the goal hold. (e) Forest and (f) ruins success counts over 20 held-out pairs per environment. In the forest, all three independently trained SAC policies are compared with the analytic baseline. In the ruins, three seed-matched adapted policies are compared on the development set, and the selected policy is evaluated on the held-out pairs.}\label{fig:supp-uav}
\end{figure*}

\begin{figure*}[!tb]
\centering\includegraphics[width=\textwidth]{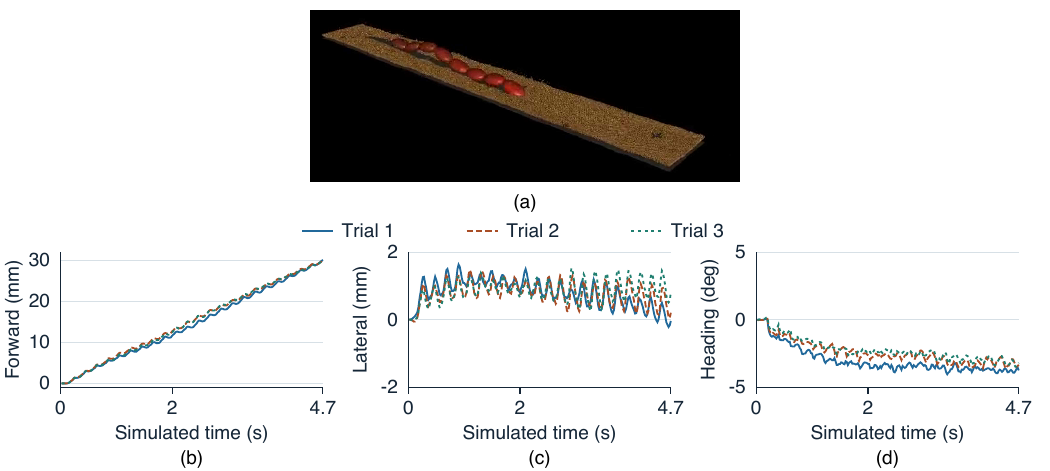}
\caption{Frozen worm policy in the training-size granular tray (main Sec.~\ref{sec:navigation-v2}). (a) Final frame of the first evaluation trial. (b) Forward displacement. (c) Lateral displacement. (d) Heading. The plots use 235 control samples per run and show all three small initial-pose perturbations of the same packing.}\label{fig:supp-worm-learning}
\end{figure*}
Table~\ref{tab:learning-protocol} summarizes the three studies in main Sec.~\ref{sec:navigation-v2} and the Go2 adaptation below. Both SAC studies use a learning rate of $3\times10^{-4}$, batch size 256, and discount factor 0.995.

\paragraph{Forest SAC.}
Soft actor-critic (SAC)~\cite{suppbib-haarnoja2018sac} learns velocity command residuals on top of analytic goal pursuit, with a frozen PPO flight controller. The curriculum starts with short routes for the first 10\% of training, then samples short, medium, and long routes with probabilities 0.2, 0.3, and 0.5. We train three independent SAC policies using different random seeds, each with 262,144 environment transitions. All three independently trained policies are frozen and evaluated on the same 20 held-out start--goal pairs as the analytic zero-residual baseline. Success requires staying within 0.15 m of the goal at no more than 0.08 m/s for 2 s, with finite states, no collision, and roll and pitch within 0.35 rad.

\paragraph{Ruins SAC adaptation.}
We conduct three adaptation runs, one for each forest SAC training seed. Each run continues training from its corresponding forest actor, critic, and optimizer, with a fresh replay buffer. Training uses 384 start--goal pairs. The three adapted policies are first compared on 24 development pairs. The best-performing policy is then frozen and evaluated against the analytic zero-residual baseline on 20 held-out ruins pairs, with matched initial states and the same goal, speed, hold, collision, and tilt guards. The low-level flight controller remains frozen throughout adaptation and evaluation.

\paragraph{Worm PPO.}
Proximal policy optimization (PPO)~\cite{suppbib-schulman2017ppo} updates six central pattern generator (CPG) parameters at 50\,Hz to generate joint angle targets. The parameters are yaw and pitch amplitudes, wave frequency, phase increment between neighboring segments, yaw-to-pitch phase offset, and yaw bias. The normalized actions are smoothed before decoding. Yaw and pitch amplitudes are limited to $\pm0.65$ rad, frequency spans 0.25--5 Hz, phase terms are limited to $\pm\pi$ rad, and yaw bias to $\pm0.15$ rad.

Training and the three deterministic evaluations use a 100\,mm-long, eight-segment worm in a flat $215\times30$\,mm XPBD tray. The bed is 3\,mm deep, with a particle diameter of 0.5\,mm. Each evaluation uses the same packing with small reset perturbations ($\pm0.2$ mm lateral and $\pm0.005$ rad yaw). The three evaluations reach the goal at 4.70 s of simulated time. A trial must advance at least 30 mm with valid physics and stay within 8 mm lateral deviation and 0.2 rad heading error.

\begin{table*}[!tb]
\caption{Training and evaluation settings for the robot learning studies.}\label{tab:learning-protocol}
\centering\small
\begin{tabular}{@{}lp{5.1cm}p{4.6cm}p{4.1cm}@{}}\toprule
Study & Observations, actions, and reward & Training and frozen components & Evaluation\\\midrule
Forest SAC & 42 local features include goal, range, robot state, and pursuit command information. Three residuals adjust velocity commands. The reward uses goal progress, local clearance and safe arrival, with collision and residual penalties. & Three independent random seeds, 262,144 transitions per seed. Frozen low-level PPO controller. & All three policies evaluated on the same 20 held-out pairs against the analytic baseline.\\
Ruins SAC & Same observation, residual action, and reward as forest. & Three seed-matched adaptations initialized from the corresponding forest policies, with 262,144 additional transitions per seed. & Three adapted policies compared on 24 development pairs. The selected policy is evaluated on 20 held-out tasks against the analytic baseline.\\
Worm PPO & Relative position and heading, body gravity and velocities, joint state, CPG phase, and action history. Six CPG actions. Reward favors forward speed and penalizes lateral deviation, heading error, and action changes, with terminal goal and failure terms. & 5,120 cumulative environment steps. Rollouts of 128 steps, batch size 64, 10 epochs, learning rate $10^{-4}$. Fixed CPG functional form and joint controller. & Three initial poses in one packing. The policy and normalization are frozen for evaluation and simulator comparison.\\
Go2 PPO & Pretrained proprioceptive observations and joint-position actions. Reward favors forward progress and reaching the exit and penalizes non-foot and proximal contacts. & One seed, 20,000 on-policy transitions and a skill-retention state library. The final actor layer and a separate critic are trained. & Three lateral starts on the training route. The pretrained and adapted policies each complete all three trials.\\\bottomrule
\end{tabular}
\end{table*}

\compactparagraph{Aerial observations and execution.}
The 42-dimensional local observation groups goal direction and range, local obstacle ranges, robot state and the analytic pursuit command. The high-level policy selects three normalized residual components at 0.5\,s intervals. The implementation adds a residual bounded by 1.5 in normalized command units and clips the combined command to $[-1,1]$ before passing it to the frozen flight controller. The evaluator records positions from the executed rotor dynamics, integrated at 240\,Hz. The flight evaluations retain their own pinned forest and ruins geometry versions. The SAC reward encourages goal progress, clearance, and safe arrival, and penalizes collisions and residual magnitude. Evaluation applies collision and attitude guards and a goal hold, as described above. Figure~\ref{fig:supp-uav} shows representative matched flights and the full held-out success counts.

\begin{figure*}[!t]
\centering\includegraphics[width=\textwidth]{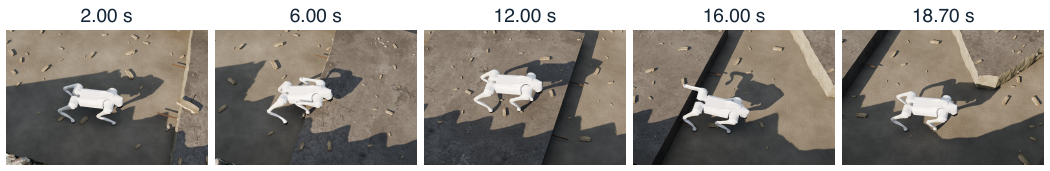}
\caption{Go2 rollout of $P_{\mathrm{adapt}}$ from the central start on the slab route. The RGB frames were rendered offline from the simulated poses. Labels show simulated time. The final frame is the terminal state of this run.}\label{fig:supp-go2-rollout}
\end{figure*}

\compactparagraph{Worm observation and control cycle.}
The PPO policy uses body-centric gravity and velocities, joint state, goal-relative displacement and heading, CPG phase and action history. The fixed CPG and joint controller translate the six parameter updates into commands every 20\,ms. Each interval comprises 200 physics steps of 0.1\,ms with five XPBD iterations per step. Training totals 5,120 environment steps including the starting policy's training. In the three evaluations, the same packing of 122,090 grains is used and only the stated pose perturbations vary. Figure~\ref{fig:supp-worm-learning} plots the three time series. Forward displacement is 30.101, 30.087 and 30.035\,mm, the mean is 30.074\,mm, maximum absolute lateral deviation is 1.633\,mm, and maximum absolute heading error is $4.025^\circ$.

\subsection{Go2 PPO adaptation in ruins}

We initialize $P_0$ from the publicly available Go2 RL Lab MoE-CTS checkpoint at iteration 82,500~\cite{suppbib-supp_go2_rllab,suppbib-supp_go2_checkpoint}. The upstream concurrent teacher--student (CTS) method uses a mixture-of-experts (MoE) student encoder. We denote the adapted policy by $P_{\mathrm{adapt}}$. The training model uses triangle-mesh signed-distance collision. The adaptation task uses a local ruins region with 76 source objects and a roughly 22--23\,cm slab obstacle. The robot ascends, traverses, and descends the slab. The task gives a 0.3\,m/s forward command with initial-heading feedback. The command's yaw rate is limited to $\pm0.3$\,rad/s.

The 45-dimensional actor observation contains angular velocity, projected gravity, the three command components, 12 joint positions, 12 joint velocities and the previous 12 actions. A frozen mixture-of-experts encoder transforms ten such observations into a 32-dimensional latent. Concatenating it with the current observation gives a 77-dimensional input to the 512--256--128 multilayer perceptron. Its 12-dimensional action mean is produced by the final affine layer, with $128\times12+12=1548$ trainable actor parameters. The encoder, hidden actor layers and action-distribution scale are frozen. PPO separately updates a fresh 49-dimensional-input critic that also observes base linear velocity and height.

At inference the policy runs at 50\,Hz. The action $a$ is clipped to $[-100,100]$ in the original policy action units and mapped to joint-angle targets in radians as $q^*=q_{\mathrm{default}}+0.25a$, where $q_{\mathrm{default}}$ is the default standing joint pose. Proportional--derivative control and physics both run at 200\,Hz with the original torque and speed limits.

Training uses 12,000 transitions in the ruins and 8,000 in a separate base-skill rehearsal environment. The PPO reward retains tracking and action-change terms and incorporates forward progress, completion, and non-foot contact penalties. A separate actor update uses the frozen $P_0$ as teacher on base-skill samples and a library of 19,200 recorded states. It minimizes the KL divergence from student to teacher action distributions and the mean squared difference of their action means. The library samples enter this retention loss independently of the on-policy budget.

\begin{figure*}[!tb]
\centering\includegraphics[width=\textwidth]{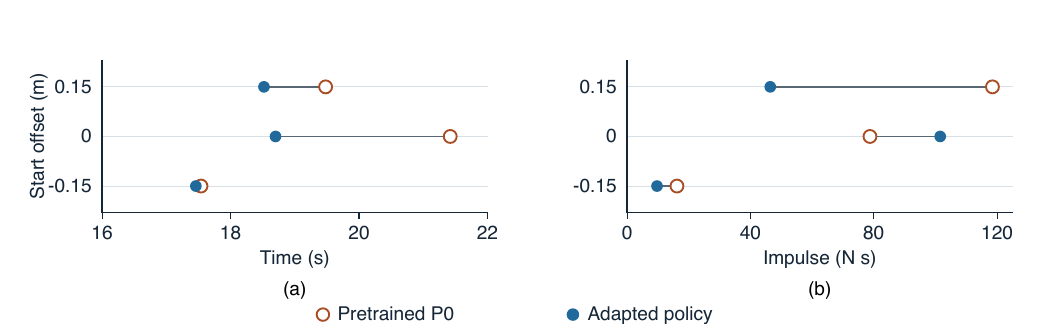}
\caption{Paired evaluation of pretrained and adapted Go2 policies. Each row represents a matched evaluation from the same lateral start offset ($-0.15$, 0, or $+0.15$\,m) along the training route. Open and filled markers denote the pretrained policy $P_0$ and adapted policy, respectively, with connecting lines indicating paired results. (a) Simulated task completion time. (b) Cumulative non-foot normal contact impulse over the route. Both policies complete all three trials.}\label{fig:supp-go2-results}
\end{figure*}

For evaluation, both policies follow the same route from three lateral offsets of $-0.15$, 0 and $+0.15$\,m. Success requires reaching the route goal, fully clearing the slab exit and staying upright and within the task region before 22\,s. Cumulative non-foot normal contact impulse is the time integral of the summed positive normal forces on non-foot robot parts. Figure~\ref{fig:supp-go2-results} compares the two policies under matched initial conditions. Across the three lateral starts, adaptation reduces the mean simulated completion time from 19.48 to 18.23\,s and the mean cumulative non-foot normal contact impulse from 71.15 to 52.57\,N\,s. Both policies complete all three trials.

\section{Adaptive simulation details}
\label{sec:supp-adaptive}
\paragraph{Excavation and local refinement.}
In the excavation comparison in main Fig.~\ref{fig:excavation}, all configurations follow the same bucket command sequence over 32 s of simulated time and use XPBD~\cite{suppbib-xpbd}. For mixed-resolution B, $\Omega_f$ is defined on the initial parent centers. We first select parents within the buffered sweep of the bucket collision hull, including the cavity, along the unloaded bucket trajectory. The buffer accounts for tracking, interpolation, parent radius, and packing displacement. We then add parents within 75 mm of those selected centers. Each selected parent is replaced with eight 25 mm particles that preserve its mass. The mixed pile starts frozen and both particle sizes activate together when the bucket approaches within the 5 mm predictive contact guard. Both remain dynamic thereafter.

\begin{table}[t]
\caption{Excavation discretization and material parameters.}\label{tab:excavation-parameters}
\centering\small
\begin{tabular}{@{}lr@{}}\toprule
Parameter & Value\\\midrule
Particle diameter (A, fine) & 25 mm\\
Particle diameter (B, mixed) & 25 and 50 mm\\
Particle diameter (C, coarse) & 50 mm\\
Particle material density & $\approx 2650$ kg/m$^3$\\
Friction coefficient & 0.7\\
Soft contact search gap & 1.25 mm\\
Soft contact relaxation & 0.3\\
Base sweep halo / tracking allowance & 0.15 / 0.8254 m\\
Packing displacement bound & 0.05 m\\
Parent center dilation in B & 0.075 m\\
Predictive activation guard in B & 0.005 m\\\bottomrule
\end{tabular}
\end{table}

\begin{table*}[!tb]
\centering\small
\caption{Initial conditions and interventions in the excavation runs. All rows use the same commanded 32\,s trajectory, 10\,ms control interval, 0.125\,ms integration step, and 80 XPBD iterations per physical step.}\label{tab:supp-excavation-configs}
\begin{tabular}{@{}lrrp{7.1cm}@{}}\toprule
Mode & Particles & Mass (kg) & Integration and intervention\\\midrule
A, fine reference & 940,376 & 20,387.680 & Dynamic from the beginning.\\
B, mixed & 657,226 & 20,387.680 & Granular motion starts at the contact trigger. Separate fine and coarse search grids with shared cross-size contacts.\\
C, coarse & 117,547 & 20,387.680 & Dynamic from the beginning.\\\bottomrule
\end{tabular}
\end{table*}

The assembly uses matching packed material provenance. Table~\ref{tab:supp-excavation-configs} lists particle counts, masses, and interventions. The two hash grids index the fine and coarse particles separately. Each particle queries both grids and accumulates same-size and cross-size contacts in one XPBD solve. The runtime and error comparisons describe the complete configurations.

\begin{figure*}[!tb]
\centering\includegraphics[width=\textwidth]{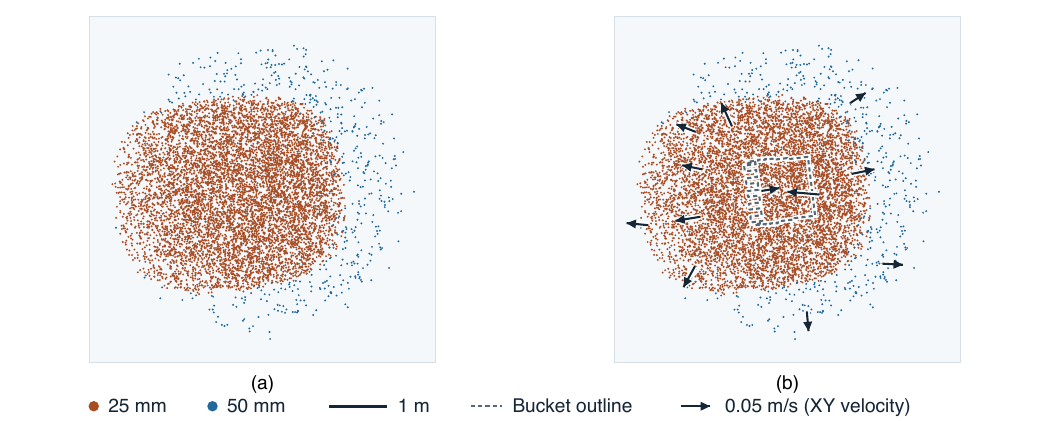}
\caption{\textbf{Excavation particle states before and after contact activation.} Matched top-down views of the same 8,000 particle IDs in configuration B from the comparison in main Fig.~\ref{fig:excavation} (Secs.~3.2--3.3). Colors indicate particle diameter. (a) Initial frozen mixed-resolution configuration at $t=0$\,s. (b) Simulation snapshot after contact activation at $t=7.70$\,s, with both particle sizes dynamic. The dashed outline shows the bucket at its recorded pose. Arrows show selected recorded horizontal particle velocities, with lengths proportional to speed.}\label{fig:supp-excavation-assembly}
\end{figure*}

The refinement region accounts for the bucket sweep, particle geometry, packing displacement, and tracking deviations. In addition to a 0.15\,m base halo, the allowance includes the 25\,mm parent-particle radius and a 0.05\,m bound on packing displacement. To account for loaded motion, we use an empirical allowance of 0.8254\,m, obtained from the maximum sampled bucket-vertex displacement (0.7754\,m) between the unloaded trajectory and a prior loaded C replay at matched times, plus a 0.05\,m margin. \RoboFieldRevision{This empirical allowance is used to select the refinement region.} Refinement is completed before the simulation. Each replaced parent contributes its mass to eight children\RoboFieldRevision{, and} the runtime uses the fixed particle identities and sizes. Figure~\ref{fig:supp-excavation-assembly} compares the frozen initial pile with the dynamic state recorded at 7.70\,s.

\compactparagraph{Load and surface metrics.}
Final and peak load sum the mass of grains inside the bucket cavity using the same membership rule across configurations. The load curve RMSE uses 3,200 aligned samples against A, and the relative final load error is $|m_B(T)-m_A(T)|/m_A(T)$. Final surface heights use a 0.1 m grid over the local region $x\in[-2,0)$ m, $y\in[-0.8,0.5)$ m. Each occupied cell uses the highest sphere top, $H=\max(0,\max_p(z_p+r_p))$, over active grains centered in the cell and outside the bucket. An empty cell is assigned the tray-floor height of zero. Table~\ref{tab:adaptive} uses the same 260 cell coordinates in this rectangle for every configuration. Figure~\ref{fig:excavation} shows the section at $y=-0.15$ m.

\paragraph{Moving window and state archive.}
The moving-window comparison in main Fig.~\ref{fig:window} uses a 10 cm worm with eight segments and 14 joints. Both simulations share the same initial particle state and frozen policy from Sec.~\ref{sec:supp-learning}. The comparison uses 3,457,440 particles in a $0.60\times0.30$\,m world over 4.70\,s. The policy was trained in the separate $0.215\times0.03$\,m tray. The initial active window is $0.30\times0.18$\,m. They use 0.5 mm grains in a 3 mm bed and compare 235 aligned trajectory samples. The initial window is $[-0.15,0.15]\times[-0.09,0.09]$ m and advances in 0.01 m increments along $+x$ at successive center-of-mass thresholds. Stability requires a speed of at most 0.002 m/s and a displacement from the stability anchor of at most 50 $\mu$m throughout a 0.20 s dwell. The selector checks stability every 0.02 s. The robot protection margin is 0.01 m. A 2 mm halo and $3\times3$ neighborhoods of 2.5 mm support tiles supply static boundaries.

\begin{figure*}[!tb]
\centering\includegraphics[width=\textwidth]{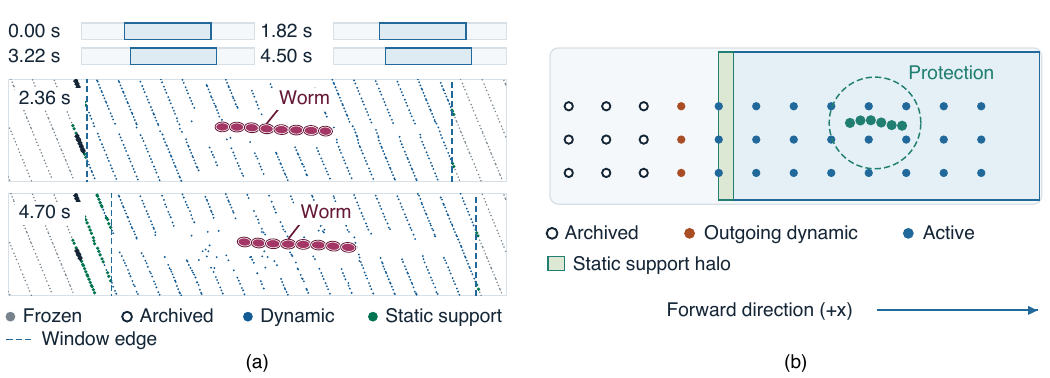}
\caption{Moving-window selection and state archiving for the simulation in main Fig.~\ref{fig:window} (Sec.~3.3). (a) Window positions at 0, 1.82, 3.22, and 4.50\,s, with cropped plan views of the worm, sampled particles, and window boundary at 2.36 and 4.70\,s. Gray dots mark particles that have yet to become dynamic. Open circles mark archived particles that retain their last simulated state outside the active solve. (b) A schematic of the trailing edge showing outgoing dynamic particles, archived particles, the protection zone and static support.}\label{fig:supp-window-archive}
\end{figure*}

Under the rule in main Eq.~\ref{eq:active-particle-set}, the initial dynamic set consists of the particles whose spheres intersect the initial window. At each control sample, the next set contains every particle whose sphere intersects the current window and any previously dynamic particle outside it that is still unstable or protected. The window advances at 1.82, 3.22, and 4.50\,s. When an outgoing particle is both stable and unprotected, its state is archived by persistent ID. Static support samples have zero effective velocity and inverse mass. If an archived particle reenters the window, the solver restores its dynamic state. Figure~\ref{fig:supp-window-archive} illustrates these states.

In a controlled test, 125 selected particles were archived and restored with zero error in the copied state. This test checks state preservation during archive and restore. The trajectory comparison in the main paper measures the error of the window simulation.

\begin{figure*}[!tb]
\centering\includegraphics[width=\textwidth]{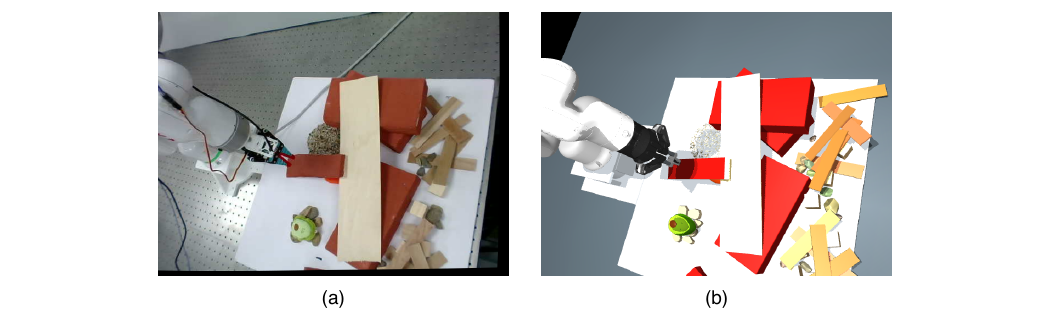}
\caption{Physical and digital workcells for the manipulation replay in main Fig.~\ref{fig:real} (Sec.~4.4). (a) Calibrated overhead view of the physical workcell. (b) Corresponding initial layout in the digital workcell.}\label{fig:supp-replay-protocol}
\end{figure*}

\begin{table*}[!tb]
\caption{Hardware and software configurations.}\label{tab:equipment}
\centering\small
\begin{tabular}{@{}p{2.8cm}p{4.0cm}p{9.4cm}@{}}\toprule
Study & Hardware & Software and execution settings\\\midrule
Scene rendering & NVIDIA RTX 3090 & Blender 5.0.1 with OptiX, $1280\times720$, 16 and 64 samples, four CPU threads.\\
Navigation planning & AMD EPYC 9754 & One numerical thread. 60~s planning budget and 150~s end-to-end wall-clock limit.\\
Forest SAC & AMD EPYC 9754 & Stable-Baselines3 v2.9.2 (SAC), 16 environments per seed, 240 Hz rotor physics.\\
Ruins SAC & AMD EPYC 9754 & Stable-Baselines3 v2.9.2 (SAC), 16 environments per seed, 240 Hz rotor physics.\\
Worm PPO & AMD EPYC 9754 (learner), NVIDIA GeForce RTX 3090 (CUDA simulation) & Stable-Baselines3 v2.9.2 (PPO) for training. Newton 1.8.0.dev0 with MuJoCo 3.16.0, Warp 1.17.0 and XPBD for simulation. Physics step 0.1 ms, control step 20 ms, five XPBD iterations.\\
Go2 PPO & CPU & RSL-RL 5.0.1 (PPO) and MuJoCo 3.13.0 with a triangle-mesh SDF plugin. Policy 50 Hz. PD and physics 200 Hz.\\
Excavation & NVIDIA L20 & Newton with MuJoCo/Warp and XPBD. Physics step 0.125 ms, 80 XPBD iterations per step.\\
Moving window & NVIDIA RTX 5090 D & XPBD, 0.1 ms physics step, five iterations per step, and 20 ms control interval.\\
Manipulation replay & MacBook Pro, Apple M2, 16 GB, macOS 14.0 & MuJoCo 3.3.7, Python 3.12.14, NumPy 2.4.1, 0.1 ms step, one physics worker. Runs use fresh serial processes, alternating mode order across pairs.\\\bottomrule
\end{tabular}
\end{table*}

\section{Real robot protocol}
\label{sec:supp-real}
\paragraph{Acquisition and labels.}
The cohort contains 64 teleoperated JAKA Mini2 trajectories across three tabletop layouts, with 20, 22, and 22 trajectories and 17/3, 18/4, and 19/3 real success/failure counts, respectively. There are 31 direct retrieval tasks and 33 two-object tasks. The target must remain visibly lifted after measured joint-feedback speed stays at or below 0.01\,rad/s for at least 0.3\,s. Clearance tasks also require the designated obstacle to be lifted before target retrieval.

\paragraph{Calibration and synchronization.}
The tabletop plane and robot base anchor the digital workcell. Object dimensions and geometry are assigned from the measured objects and their replay collision assets. Initial instance poses are aligned to the recorded layout and checked against the plane and required supports. Object masses and inertias come from the reconstructed case models, which use nominal masses for objects without measured mass. Contact coefficients come from the scene and case contact files, which include material-based estimates. The fixed overhead and external cameras have separate intrinsic and extrinsic calibrations, and frames are aligned to the measured motion using host timestamps. Figure~\ref{fig:supp-replay-protocol} compares the physical workcell and the corresponding digital initial layout.

\paragraph{Paired replay protocol.}
Full and selective activation start from the same scene and support geometry. In both modes, bounded joint servo control tracks the same measured joint feedback and replays the same binary gripper commands. Both use the same physical parameters, fixed in the model package before timing, and share the stopping criterion, simulated duration, and step count.

\textbf{Task-specific object selection.}
The Physics Configurator described in main Sec.~3.2 uses the shared scene record, robot geometry, and task specification to propose \RoboFieldRevision{task-relevant} objects and physical representations based on task stages and object relations. Deterministic processing resolves entity IDs, includes objects that can enter the robot or carried-object swept envelopes, and recursively adds movable supports. \RoboFieldRevision{The resulting activation set is compiled before replay and remains fixed throughout execution.} Model implementation details are provided in Sec.~\ref{sec:supp-repro}.

Full activation simulates all movable objects dynamically. Selective activation \RoboFieldRevision{dynamically} simulates the \RoboFieldRevision{task-relevant} objects \RoboFieldRevision{identified by the Physics Configurator, together with their} movable supports. The remaining objects retain static collision geometry. Task stages govern the lift sequence and attachment handling within the selected set.

\paragraph{Comparison.}
Selective activation reduces total replay runtime by 15.0\% across the 64 paired trajectories. Both modes agree with real-world success/failure labels on 63 of 64 trajectories (98.44\%; Table~\ref{tab:real}). The only mismatch occurs in a green-target retrieval task in layout 2, where the simulated target rises by at most 2.06\,mm in either mode, below the 20\,mm success threshold. This evaluation measures task-level outcomes rather than the accuracy of individual object trajectories, contact forces, or deformation.

\section{Implementation and reproducibility}
\label{sec:supp-repro}
\label{sec:implementation}

\paragraph{Model implementations.}
The Scene Planner, Scene Reviewer, and Physics Configurator use GPT-6 Astra via ChatGPT. For on-demand asset generation, the Asset Designer uses DeepSeek V4.1 Flash via Volcengine Ark~\cite{suppbib-supp_deepseek_ark}, the Visual Designer uses Doubao Seedream 5.0 Pro~\cite{suppbib-supp_seedream}, and the 3D Modeler uses Tencent Hunyuan3D Pro 3.1~\cite{suppbib-supp_hunyuan3d}.

Table~\ref{tab:equipment} lists the hardware and software. Runtime denotes elapsed execution time under the stated measurement boundary. Simulated time denotes the duration represented by the model. Sec.~\ref{sec:supp-nav} specifies the two planning timing definitions.

\paragraph{Rendering settings.}
Table~\ref{tab:equipment} reports the rendering configuration used for scene visualization. L4 scenes are rendered at $1280\times720$ using the native camera and the same renderer. We use 16- and 64-sample settings and visually inspect the first 64-sample render of each scene to check its visible content.

\paragraph{Excavation timing.}
Runtime includes initialization, simulation, logging, snapshots, and archive output. Particle preparation, offline checks, and rendering are outside the timing boundary. All three configurations are run once on L20 hosts.

\paragraph{Moving-window timing.}
Runtime includes physics, boundary construction, and particle selection and excludes common initial-state preparation.

\paragraph{Manipulation replay timing.}
Timing starts at replay-package load and ends after the natural stopping point and final log write. It includes initialization, physics, control, state migration, and I/O. Model and activation-set construction, offline copying, checks, and rendering are outside the timing boundary.

\par
\FloatBarrier
{\small
\setlength{\bibsep}{1pt plus 1pt minus 1pt}
\interlinepenalty=10000

}


\begin{thebibliography}{41}
\providecommand{\natexlab}[1]{#1}
\providecommand{\url}[1]{\texttt{#1}}
\expandafter\ifx\csname urlstyle\endcsname\relax
  \providecommand{\doi}[1]{doi: #1}\else
  \providecommand{\doi}{doi: \begingroup \urlstyle{rm}\Url}\fi

\bibitem[Adams et~al.(2007)Adams, Pauly, Keiser, and Guibas]{adams2007adaptive}
Bart Adams, Mark Pauly, Richard Keiser, and Leonidas~J. Guibas.
\newblock Adaptively sampled particle fluids.
\newblock \emph{ACM Transactions on Graphics}, 26\penalty0 (3):\penalty0 48:1--48:7, 2007.

\bibitem[Cabon et~al.(2020)Cabon, Murray, and Humenberger]{cabon2020virtualkitti2}
Yohann Cabon, Naila Murray, and Martin Humenberger.
\newblock Virtual {KITTI} 2.
\newblock \emph{arXiv preprint arXiv:2001.10773}, 2020.

\bibitem[Deitke et~al.(2022)Deitke, VanderBilt, Herrasti, Weihs, Salvador, Ehsani, Han, Kolve, Farhadi, Kembhavi, and Mottaghi]{procthor}
Matt Deitke, Eli VanderBilt, Alvaro Herrasti, Luca Weihs, Jordi Salvador, Kiana Ehsani, Winson Han, Eric Kolve, Ali Farhadi, Aniruddha Kembhavi, and Roozbeh Mottaghi.
\newblock {ProcTHOR}: Large-scale embodied {AI} using procedural generation.
\newblock In \emph{Advances in Neural Information Processing Systems}, pages 5982--5994, 2022.

\bibitem[Deitke et~al.(2023{\natexlab{a}})Deitke, Liu, Wallingford, Ngo, Michel, Kusupati, Fan, Laforte, Voleti, Gadre, VanderBilt, Kembhavi, Vondrick, Gkioxari, Ehsani, Schmidt, and Farhadi]{objaversexl}
Matt Deitke, Ruoshi Liu, Matthew Wallingford, Huong Ngo, Oscar Michel, Aditya Kusupati, Alan Fan, Christian Laforte, Vikram Voleti, Samir~Yitzhak Gadre, Eli VanderBilt, Aniruddha Kembhavi, Carl Vondrick, Georgia Gkioxari, Kiana Ehsani, Ludwig Schmidt, and Ali Farhadi.
\newblock {Objaverse-XL}: A universe of {10M+} {3D} objects.
\newblock In \emph{Advances in Neural Information Processing Systems}, pages 35799--35813, 2023{\natexlab{a}}.

\bibitem[Deitke et~al.(2023{\natexlab{b}})Deitke, Schwenk, Salvador, Weihs, Michel, VanderBilt, Schmidt, Ehsani, Kembhavi, and Farhadi]{objaverse}
Matt Deitke, Dustin Schwenk, Jordi Salvador, Luca Weihs, Oscar Michel, Eli VanderBilt, Ludwig Schmidt, Kiana Ehsani, Aniruddha Kembhavi, and Ali Farhadi.
\newblock {Objaverse}: A universe of annotated {3D} objects.
\newblock In \emph{Proceedings of the IEEE/CVF Conference on Computer Vision and Pattern Recognition}, pages 13142--13153, 2023{\natexlab{b}}.

\bibitem[Feng et~al.(2023)Feng, Zhu, Fu, Jampani, Akula, He, Basu, Wang, and Wang]{layoutgpt}
Weixi Feng, Wanrong Zhu, Tsu-Jui Fu, Varun Jampani, Arjun Akula, Xuehai He, Sugato Basu, Xin~Eric Wang, and William~Yang Wang.
\newblock {LayoutGPT}: Compositional visual planning and generation with large language models.
\newblock In \emph{Advances in Neural Information Processing Systems}, pages 18225--18250, 2023.

\bibitem[Fu et~al.(2021{\natexlab{a}})Fu, Cai, Gao, Zhang, Wang, Li, Zeng, Sun, Jia, Zhao, and Zhang]{threedfront}
Huan Fu, Bowen Cai, Lin Gao, Ling-Xiao Zhang, Jiaming Wang, Cao Li, Qixun Zeng, Chengyue Sun, Rongfei Jia, Binqiang Zhao, and Hao Zhang.
\newblock {3D-FRONT}: {3D} furnished rooms with {layOuts} and {semaNTics}.
\newblock In \emph{Proceedings of the IEEE/CVF International Conference on Computer Vision}, pages 10933--10942, 2021{\natexlab{a}}.

\bibitem[Fu et~al.(2021{\natexlab{b}})Fu, Jia, Gao, Gong, Zhao, Maybank, and Tao]{threedfuture}
Huan Fu, Rongfei Jia, Lin Gao, Mingming Gong, Binqiang Zhao, Steve Maybank, and Dacheng Tao.
\newblock {3D-FUTURE}: {3D} furniture shape with {TextURE}.
\newblock \emph{International Journal of Computer Vision}, 129:\penalty0 3313--3337, 2021{\natexlab{b}}.

\bibitem[Gu et~al.(2023)Gu, Xiang, Li, Ling, Liu, Mu, Tang, Tao, Wei, Yao, Yuan, Xie, Huang, Chen, and Su]{maniskill2023}
Jiayuan Gu, Fanbo Xiang, Xuanlin Li, Zhan Ling, Xiqiang Liu, Tongzhou Mu, Yihe Tang, Stone Tao, Xinyue Wei, Yunchao Yao, Xiaodi Yuan, Pengwei Xie, Zhiao Huang, Rui Chen, and Hao Su.
\newblock {ManiSkill2}: A unified benchmark for generalizable manipulation skills.
\newblock In \emph{International Conference on Learning Representations}, 2023.

\bibitem[Haarnoja et~al.(2018)Haarnoja, Zhou, Abbeel, and Levine]{haarnoja2018sac}
Tuomas Haarnoja, Aurick Zhou, Pieter Abbeel, and Sergey Levine.
\newblock Soft actor-critic: Off-policy maximum entropy deep reinforcement learning with a stochastic actor.
\newblock In \emph{Proceedings of the 35th International Conference on Machine Learning}, pages 1861--1870, 2018.

\bibitem[Hart et~al.(1968)Hart, Nilsson, and Raphael]{hart1968astar}
Peter~E. Hart, Nils~J. Nilsson, and Bertram Raphael.
\newblock A formal basis for the heuristic determination of minimum cost paths.
\newblock \emph{IEEE Transactions on Systems Science and Cybernetics}, 4\penalty0 (2):\penalty0 100--107, 1968.

\bibitem[H{\"o}llein et~al.(2023)H{\"o}llein, Cao, Owens, Johnson, and Nie{\ss}ner]{text2room}
Lukas H{\"o}llein, Ang Cao, Andrew Owens, Justin Johnson, and Matthias Nie{\ss}ner.
\newblock {Text2Room}: Extracting textured {3D} meshes from {2D} text-to-image models.
\newblock In \emph{Proceedings of the IEEE/CVF International Conference on Computer Vision}, pages 7909--7920, 2023.

\bibitem[Jiang et~al.(2015)Jiang, Schroeder, Selle, Teran, and Stomakhin]{jiang2015apic}
Chenfanfu Jiang, Craig Schroeder, Andrew Selle, Joseph Teran, and Alexey Stomakhin.
\newblock The affine particle-in-cell method.
\newblock \emph{ACM Transactions on Graphics}, 34\penalty0 (4):\penalty0 51:1--51:10, 2015.

\bibitem[Karaman and Frazzoli(2011)]{karaman2011rrtstar}
Sertac Karaman and Emilio Frazzoli.
\newblock Sampling-based algorithms for optimal motion planning.
\newblock \emph{The International Journal of Robotics Research}, 30\penalty0 (7):\penalty0 846--894, 2011.

\bibitem[Li et~al.(2022)Li, Xia, Mart{\'i}n-Mart{\'i}n, Lingelbach, Srivastava, Shen, Vainio, Gokmen, Dharan, Jain, Kurenkov, Liu, Gweon, Wu, Fei-Fei, and Savarese]{igibson2022}
Chengshu Li, Fei Xia, Roberto Mart{\'i}n-Mart{\'i}n, Michael Lingelbach, Sanjana Srivastava, Bokui Shen, Kent~Elliott Vainio, Cem Gokmen, Gokul Dharan, Tanish Jain, Andrey Kurenkov, Karen Liu, Hyowon Gweon, Jiajun Wu, Li Fei-Fei, and Silvio Savarese.
\newblock {iGibson 2.0}: Object-centric simulation for robot learning of everyday household tasks.
\newblock In \emph{Proceedings of the 5th Conference on Robot Learning}, pages 455--465. PMLR, 2022.

\bibitem[Li et~al.(2023)Li, Zhang, Wong, Gokmen, Srivastava, Mart{\'i}n-Mart{\'i}n, Wang, Levine, Lingelbach, Sun, Anvari, Hwang, Sharma, Aydin, Bansal, Hunter, Kim, Lou, Matthews, Villa-Renteria, Tang, Tang, Xia, Savarese, Gweon, Liu, Wu, and Fei-Fei]{behavior1k2023}
Chengshu Li, Ruohan Zhang, Josiah Wong, Cem Gokmen, Sanjana Srivastava, Roberto Mart{\'i}n-Mart{\'i}n, Chen Wang, Gabrael Levine, Michael Lingelbach, Jiankai Sun, Mona Anvari, Minjune Hwang, Manasi Sharma, Arman Aydin, Dhruva Bansal, Samuel Hunter, Kyu-Young Kim, Alan Lou, Caleb~R. Matthews, Ivan Villa-Renteria, Jerry~Huayang Tang, Claire Tang, Fei Xia, Silvio Savarese, Hyowon Gweon, Karen Liu, Jiajun Wu, and Li Fei-Fei.
\newblock {BEHAVIOR-1K}: A benchmark for embodied {AI} with 1,000 everyday activities and realistic simulation.
\newblock In \emph{Proceedings of the 6th Conference on Robot Learning}, pages 80--93. PMLR, 2023.

\bibitem[Likhachev et~al.(2003)Likhachev, Gordon, and Thrun]{likhachev2003ara}
Maxim Likhachev, Geoffrey~J. Gordon, and Sebastian Thrun.
\newblock {ARA*}: Anytime {A*} with provable bounds on sub-optimality.
\newblock In \emph{Advances in Neural Information Processing Systems}, pages 767--774, 2003.

\bibitem[Macklin et~al.(2016)Macklin, M{\"u}ller, and Chentanez]{xpbd}
Miles Macklin, Matthias M{\"u}ller, and Nuttapong Chentanez.
\newblock {XPBD}: Position-based simulation of compliant constrained dynamics.
\newblock In \emph{Proceedings of the 9th International Conference on Motion in Games}, pages 49--54, 2016.

\bibitem[Makoviychuk et~al.(2021)Makoviychuk, Wawrzyniak, Guo, Lu, Storey, Macklin, Hoeller, Rudin, Allshire, Handa, and State]{isaacgym2021}
Viktor Makoviychuk, Lukasz Wawrzyniak, Yunrong Guo, Michelle Lu, Kier Storey, Miles Macklin, David Hoeller, Nikita Rudin, Arthur Allshire, Ankur Handa, and Gavriel State.
\newblock {Isaac Gym}: High performance {GPU} based physics simulation for robot learning.
\newblock In \emph{Proceedings of the Neural Information Processing Systems Track on Datasets and Benchmarks}, 2021.

\bibitem[{MuJoCo Developers}(2025)]{mujoco_sleeping340}
{MuJoCo Developers}.
\newblock {MuJoCo 3.4.0 documentation: Sleeping islands}.
\newblock \url{https://mujoco.readthedocs.io/en/3.4.0/computation/index.html}, 2025.

\bibitem[M{\"u}ller et~al.(2007)M{\"u}ller, Heidelberger, Hennix, and Ratcliff]{muller2007pbd}
Matthias M{\"u}ller, Bruno Heidelberger, Marcus Hennix, and John Ratcliff.
\newblock Position based dynamics.
\newblock \emph{Journal of Visual Communication and Image Representation}, 18\penalty0 (2):\penalty0 109--118, 2007.

\bibitem[Nasiriany et~al.(2024)Nasiriany, Maddukuri, Zhang, Parikh, Lo, Joshi, Mandlekar, and Zhu]{robocasa}
Soroush Nasiriany, Abhiram Maddukuri, Lance Zhang, Adeet Parikh, Aaron Lo, Abhishek Joshi, Ajay Mandlekar, and Yuke Zhu.
\newblock {RoboCasa}: Large-scale simulation of household tasks for generalist robots.
\newblock In \emph{Robotics: Science and Systems}, 2024.

\bibitem[Niu et~al.(2026)Niu, Xie, Peng, Zhao, and Liu]{niu2026skelwam}
Pengjun Niu, Yujia Xie, Rui Peng, Hang Zhao, and Ke Liu.
\newblock {SkelWAM}: A skeleton-guided world-action model for zero-shot cross-embodiment manipulation.
\newblock \emph{arXiv preprint arXiv:2609.21983}, 2026.

\bibitem[Paschalidou et~al.(2021)Paschalidou, Kar, Shugrina, Kreis, Geiger, and Fidler]{atiss}
Despoina Paschalidou, Amlan Kar, Maria Shugrina, Karsten Kreis, Andreas Geiger, and Sanja Fidler.
\newblock {ATISS}: Autoregressive transformers for indoor scene synthesis.
\newblock In \emph{Advances in Neural Information Processing Systems}, pages 12013--12026, 2021.

\bibitem[Raistrick et~al.(2023)Raistrick, Lipson, Ma, Mei, Wang, Zuo, Kayan, Wen, Han, Wang, Newell, Law, Goyal, Yang, and Deng]{infinigen}
Alexander Raistrick, Lahav Lipson, Zeyu Ma, Lingjie Mei, Mingzhe Wang, Yiming Zuo, Karhan Kayan, Hongyu Wen, Beining Han, Yihan Wang, Alejandro Newell, Hei Law, Ankit Goyal, Kaiyu Yang, and Jia Deng.
\newblock Infinite photorealistic worlds using procedural generation.
\newblock In \emph{Proceedings of the IEEE/CVF Conference on Computer Vision and Pattern Recognition}, pages 12630--12641, 2023.

\bibitem[Raistrick et~al.(2024)Raistrick, Mei, Kayan, Yan, Zuo, Han, Wen, Parakh, Alexandropoulos, Lipson, Ma, and Deng]{infinigen_indoors}
Alexander Raistrick, Lingjie Mei, Karhan Kayan, David Yan, Yiming Zuo, Beining Han, Hongyu Wen, Meenal Parakh, Stamatis Alexandropoulos, Lahav Lipson, Zeyu Ma, and Jia Deng.
\newblock {Infinigen Indoors}: Photorealistic indoor scenes using procedural generation.
\newblock In \emph{Proceedings of the IEEE/CVF Conference on Computer Vision and Pattern Recognition}, pages 21783--21794, 2024.

\bibitem[Redon et~al.(2005)Redon, Galoppo, and Lin]{redon2005}
St{\'e}phane Redon, Nico Galoppo, and Ming~C. Lin.
\newblock Adaptive dynamics of articulated bodies.
\newblock \emph{ACM Transactions on Graphics}, 24\penalty0 (3):\penalty0 936--945, 2005.

\bibitem[Ros et~al.(2016)Ros, Sellart, Materzynska, Vazquez, and Lopez]{ros2016synthia}
German Ros, Laura Sellart, Joanna Materzynska, David Vazquez, and Antonio~M. Lopez.
\newblock The {SYNTHIA} dataset: A large collection of synthetic images for semantic segmentation of urban scenes.
\newblock In \emph{Proceedings of the IEEE Conference on Computer Vision and Pattern Recognition}, pages 3234--3243, 2016.

\bibitem[Rudin et~al.(2022)Rudin, Hoeller, Reist, and Hutter]{rudin2022learningtowalk}
Nikita Rudin, David Hoeller, Philipp Reist, and Marco Hutter.
\newblock Learning to walk in minutes using massively parallel deep reinforcement learning.
\newblock In \emph{Proceedings of the 5th Conference on Robot Learning}, pages 91--100. PMLR, 2022.

\bibitem[Savva et~al.(2019)Savva, Kadian, Maksymets, Zhao, Wijmans, Jain, Straub, Liu, Koltun, Malik, Parikh, and Batra]{habitat2019}
Manolis Savva, Abhishek Kadian, Oleksandr Maksymets, Yili Zhao, Erik Wijmans, Bhavana Jain, Julian Straub, Jia Liu, Vladlen Koltun, Jitendra Malik, Devi Parikh, and Dhruv Batra.
\newblock {Habitat}: A platform for embodied {AI} research.
\newblock In \emph{Proceedings of the IEEE/CVF International Conference on Computer Vision}, pages 9339--9347, 2019.

\bibitem[Schulman et~al.(2017)Schulman, Wolski, Dhariwal, Radford, and Klimov]{schulman2017ppo}
John Schulman, Filip Wolski, Prafulla Dhariwal, Alec Radford, and Oleg Klimov.
\newblock Proximal policy optimization algorithms.
\newblock \emph{arXiv preprint arXiv:1707.06347}, 2017.

\bibitem[Stomakhin et~al.(2013)Stomakhin, Schroeder, Chai, Teran, and Selle]{stomakhin2013snow}
Alexey Stomakhin, Craig Schroeder, Lawrence Chai, Joseph Teran, and Andrew Selle.
\newblock A material point method for snow simulation.
\newblock \emph{ACM Transactions on Graphics}, 32\penalty0 (4):\penalty0 102:1--102:10, 2013.

\bibitem[Tang et~al.(2024)Tang, Nie, Markhasin, Dai, Thies, and Nie{\ss}ner]{diffuscene}
Jiapeng Tang, Yinyu Nie, Lev Markhasin, Angela Dai, Justus Thies, and Matthias Nie{\ss}ner.
\newblock {DiffuScene}: Denoising diffusion models for generative indoor scene synthesis.
\newblock In \emph{Proceedings of the IEEE/CVF Conference on Computer Vision and Pattern Recognition}, pages 20507--20518, 2024.

\bibitem[Todorov et~al.(2012)Todorov, Erez, and Tassa]{mujoco}
Emanuel Todorov, Tom Erez, and Yuval Tassa.
\newblock {MuJoCo}: A physics engine for model-based control.
\newblock In \emph{IEEE/RSJ International Conference on Intelligent Robots and Systems}, pages 5026--5033, 2012.

\bibitem[Wang et~al.(2021)Wang, Yeshwanth, and Nie{\ss}ner]{sceneformer}
Xinpeng Wang, Chandan Yeshwanth, and Matthias Nie{\ss}ner.
\newblock {SceneFormer}: Indoor scene generation with transformers.
\newblock In \emph{International Conference on 3D Vision}, pages 106--115, 2021.

\bibitem[Wang et~al.(2024)Wang, Xian, Chen, Wang, Wang, Fragkiadaki, Erickson, Held, and Gan]{robogen}
Yufei Wang, Zhou Xian, Feng Chen, Tsun-Hsuan Wang, Yian Wang, Katerina Fragkiadaki, Zackory Erickson, David Held, and Chuang Gan.
\newblock {RoboGen}: Towards unleashing infinite data for automated robot learning via generative simulation.
\newblock In \emph{Proceedings of the 41st International Conference on Machine Learning}, pages 51936--51983, 2024.

\bibitem[Xia et~al.(2018)Xia, Zamir, He, Sax, Malik, and Savarese]{gibson2018}
Fei Xia, Amir~R. Zamir, Zhiyang He, Alexander Sax, Jitendra Malik, and Silvio Savarese.
\newblock {Gibson Env}: Real-world perception for embodied agents.
\newblock In \emph{Proceedings of the IEEE Conference on Computer Vision and Pattern Recognition}, pages 9068--9079, 2018.

\bibitem[Xia et~al.(2026)Xia, Li, Li, Ma, Xu, Liu, Cui, Lin, Ma, Wang, Song, and Wei]{sage2026}
Hongchi Xia, Xuan Li, Zhaoshuo Li, Qianli Ma, Jiashu Xu, Ming-Yu Liu, Yin Cui, Tsung-Yi Lin, Wei-Chiu Ma, Shenlong Wang, Shuran Song, and Fangyin Wei.
\newblock {SAGE}: Scalable agentic {3D} scene generation for embodied {AI}.
\newblock In \emph{Proceedings of the IEEE/CVF Conference on Computer Vision and Pattern Recognition}, pages 22358--22368, 2026.

\bibitem[Xiang et~al.(2020)Xiang, Qin, Mo, Xia, Zhu, Liu, Liu, Jiang, Yuan, Wang, Yi, Chang, Guibas, and Su]{sapien2020}
Fanbo Xiang, Yuzhe Qin, Kaichun Mo, Yikuan Xia, Hao Zhu, Fangchen Liu, Minghua Liu, Hanxiao Jiang, Yifu Yuan, He Wang, Li Yi, Angel~X. Chang, Leonidas~J. Guibas, and Hao Su.
\newblock {SAPIEN}: A {SimulAted} part-based interactive {ENvironment}.
\newblock In \emph{Proceedings of the IEEE/CVF Conference on Computer Vision and Pattern Recognition}, pages 11097--11107, 2020.

\bibitem[Yang et~al.(2024{\natexlab{a}})Yang, Jia, Zhi, and Huang]{physcene}
Yandan Yang, Baoxiong Jia, Peiyuan Zhi, and Siyuan Huang.
\newblock {PhyScene}: Physically interactable {3D} scene synthesis for embodied {AI}.
\newblock In \emph{Proceedings of the IEEE/CVF Conference on Computer Vision and Pattern Recognition}, pages 16262--16272, 2024{\natexlab{a}}.

\bibitem[Yang et~al.(2024{\natexlab{b}})Yang, Sun, Weihs, VanderBilt, Herrasti, Han, Wu, Haber, Krishna, Liu, Callison-Burch, Yatskar, Kembhavi, and Clark]{holodeck}
Yue Yang, Fan-Yun Sun, Luca Weihs, Eli VanderBilt, Alvaro Herrasti, Winson Han, Jiajun Wu, Nick Haber, Ranjay Krishna, Lingjie Liu, Chris Callison-Burch, Mark Yatskar, Aniruddha Kembhavi, and Christopher Clark.
\newblock {Holodeck}: Language guided generation of {3D} embodied {AI} environments.
\newblock In \emph{Proceedings of the IEEE/CVF Conference on Computer Vision and Pattern Recognition}, pages 16227--16237, 2024{\natexlab{b}}.

\end{thebibliography}

\begin{thebibliography}{21}
\providecommand{\natexlab}[1]{#1}
\providecommand{\url}[1]{\texttt{#1}}
\expandafter\ifx\csname urlstyle\endcsname\relax
  \providecommand{\doi}[1]{doi: #1}\else
  \providecommand{\doi}{doi: \begingroup \urlstyle{rm}\Url}\fi

\bibitem[Bender and Koschier(2017)]{suppbib-bender2017dfsph}
Jan Bender and Dan Koschier.
\newblock Divergence-free {SPH} for incompressible and viscous fluids.
\newblock \emph{IEEE Transactions on Visualization and Computer Graphics},
  23\penalty0 (3):\penalty0 1193--1206, 2017.

\bibitem[Dadvand et~al.(2010)Dadvand, Rossi, and O{\~n}ate]{suppbib-dadvand2010kratos}
Pooyan Dadvand, Riccardo Rossi, and Eugenio O{\~n}ate.
\newblock An object-oriented environment for developing finite element codes
  for multi-disciplinary applications.
\newblock \emph{Archives of Computational Methods in Engineering}, 17:\penalty0
  253--297, 2010.

\bibitem[Faure et~al.(2012)Faure, Duriez, Delingette, Allard, Gilles,
  Marchesseau, Talbot, Courtecuisse, Bousquet, Peterlik, and
  Cotin]{suppbib-faure2012sofa}
Fran{\c c}ois Faure, Christian Duriez, Herv{\'e} Delingette, J{\'e}r{\'e}mie
  Allard, Benjamin Gilles, St{\'e}phanie Marchesseau, Hugo Talbot, Hadrien
  Courtecuisse, Guillaume Bousquet, Igor Peterlik, and St{\'e}phane Cotin.
\newblock {SOFA}: A multi-model framework for interactive physical simulation.
\newblock In \emph{Soft Tissue Biomechanical Modeling for Computer Assisted
  Surgery}, pages 283--321. Springer, 2012.

\bibitem[{Federal Office of Topography swisstopo}(2024)]{suppbib-supp_swisstopo}
{Federal Office of Topography swisstopo}.
\newblock {\RoboFieldRevision{swissALTI3D and SWISSIMAGE 10 cm}}.
\newblock \RoboFieldRevision{Official geodata product documentation}, 2024.
\newblock \RoboFieldRevision{Rh\^one Glacier area data include swissALTI3D
  elevation tiles (2024) and SWISSIMAGE orthophoto tiles (2023)}.

\bibitem[{Gazebo Contributors}(n.d.)]{suppbib-gazebo_hydrodynamics}
{Gazebo Contributors}.
\newblock {Gazebo Sim}: Underwater vehicles.
\newblock \RoboFieldRevision{Gazebo Sim 8 documentation}, n.d.
\newblock Accessed November 9, 2026.

\bibitem[Haarnoja et~al.(2018)Haarnoja, Zhou, Abbeel, and
  Levine]{suppbib-haarnoja2018sac}
Tuomas Haarnoja, Aurick Zhou, Pieter Abbeel, and Sergey Levine.
\newblock Soft actor-critic: Off-policy maximum entropy deep reinforcement
  learning with a stochastic actor.
\newblock In \emph{Proceedings of the 35th International Conference on Machine
  Learning}, pages 1861--1870, 2018.

\bibitem[Lee et~al.(2018)Lee, Grey, Ha, Kunz, Jain, Ye, Srinivasa, Stilman, and
  Liu]{suppbib-lee2018dart}
Jeongseok Lee, Michael~X. Grey, Sehoon Ha, Tobias Kunz, Sumit Jain, Yuting Ye,
  Siddhartha~S. Srinivasa, Mike Stilman, and C.~Karen Liu.
\newblock {DART}: Dynamic animation and robotics toolkit.
\newblock \emph{Journal of Open Source Software}, 3\penalty0 (22):\penalty0
  500, 2018.

\bibitem[Macklin et~al.(2016)Macklin, M{\"u}ller, and Chentanez]{suppbib-xpbd}
Miles Macklin, Matthias M{\"u}ller, and Nuttapong Chentanez.
\newblock {XPBD}: Position-based simulation of compliant constrained dynamics.
\newblock In \emph{Proceedings of the 9th International Conference on Motion in
  Games}, pages 49--54, 2016.

\bibitem[Mark(2024)]{suppbib-supp_blendkit}
Japhet Mark.
\newblock {Concrete \RoboFieldRevision{Debris}}.
\newblock \RoboFieldRevision{BlenderKit 3D asset}, 2024.
\newblock Royalty Free license.

\bibitem[{Poly Haven}(n.d.)]{suppbib-supp_polyhaven}
{Poly Haven}.
\newblock {\RoboFieldRevision{Asset License}}.
\newblock \RoboFieldRevision{Official asset library documentation}, n.d.
\newblock CC0 1.0 Universal.

\bibitem[Raistrick et~al.(2023)Raistrick, Lipson, Ma, Mei, Wang, Zuo, Kayan,
  Wen, Han, Wang, Newell, Law, Goyal, Yang, and Deng]{suppbib-infinigen}
Alexander Raistrick, Lahav Lipson, Zeyu Ma, Lingjie Mei, Mingzhe Wang, Yiming
  Zuo, Karhan Kayan, Hongyu Wen, Beining Han, Yihan Wang, Alejandro Newell, Hei
  Law, Ankit Goyal, Kaiyu Yang, and Jia Deng.
\newblock Infinite photorealistic worlds using procedural generation.
\newblock In \emph{Proceedings of the IEEE/CVF Conference on Computer Vision
  and Pattern Recognition}, pages 12630--12641, 2023.

\bibitem[Schulman et~al.(2017)Schulman, Wolski, Dhariwal, Radford, and
  Klimov]{suppbib-schulman2017ppo}
John Schulman, Filip Wolski, Prafulla Dhariwal, Alec Radford, and Oleg Klimov.
\newblock Proximal policy optimization algorithms.
\newblock \emph{arXiv preprint arXiv:1707.06347}, 2017.

\bibitem[{SPlisHSPlasH Contributors}(n.d.)]{suppbib-splishsplash}
{SPlisHSPlasH Contributors}.
\newblock {SPlisHSPlasH}: Documentation.
\newblock \RoboFieldRevision{Software documentation on Read the Docs}, n.d.
\newblock Accessed November 9, 2026.

\bibitem[Tasora et~al.(2016)Tasora, Serban, Mazhar, Pazouki, Melanz,
  Fleischmann, Taylor, Sugiyama, and Negrut]{suppbib-tasora2016chrono}
Alessandro Tasora, Radu Serban, Hammad Mazhar, Arman Pazouki, Daniel Melanz,
  Jonathan Fleischmann, Michael Taylor, Hiroyuki Sugiyama, and Dan Negrut.
\newblock Chrono: An open source multi-physics dynamics engine.
\newblock In \emph{High Performance Computing in Science and Engineering},
  pages 19--49. Springer, 2016.

\bibitem[{Tencent Cloud}(n.d.)]{suppbib-supp_hunyuan3d}
{Tencent Cloud}.
\newblock {\RoboFieldRevision{Hunyuan3D Pro API}}.
\newblock \RoboFieldRevision{Official API documentation}, n.d.
\newblock \RoboFieldRevision{Model version 3.1, endpoint
  \mbox{SubmitHunyuanTo3DProJob}}.

\bibitem[{The Newton Contributors}(2025)]{suppbib-newton}
{The Newton Contributors}.
\newblock {Newton}: {GPU}-accelerated physics simulation for robotics and
  simulation research.
\newblock GitHub software repository, 2025.
\newblock Accessed November 9, 2026.

\bibitem[Todorov et~al.(2012)Todorov, Erez, and Tassa]{suppbib-mujoco}
Emanuel Todorov, Tom Erez, and Yuval Tassa.
\newblock {MuJoCo}: A physics engine for model-based control.
\newblock In \emph{IEEE/RSJ International Conference on Intelligent Robots and
  Systems}, pages 5026--5033, 2012.

\bibitem[{Volcengine}(n.d{\natexlab{a}})]{suppbib-supp_deepseek_ark}
{Volcengine}.
\newblock {{{Volcengine Ark}: Model list}}.
\newblock Volcengine Ark documentation, n.d{\natexlab{a}}.

\bibitem[{Volcengine}(n.d{\natexlab{b}})]{suppbib-supp_seedream}
{Volcengine}.
\newblock {{{Doubao Seedream 5.0 Pro}}}.
\newblock Volcengine Ark documentation, n.d{\natexlab{b}}.

\bibitem[Wu(2026)]{suppbib-supp_go2_checkpoint}
Tianyang Wu.
\newblock {\RoboFieldRevision{Go2 RL Data}}.
\newblock Hugging Face model repository
  \RoboFieldRevision{\texttt{wty-yy/go2\_rl\_gym\_data}}, 2026.
\newblock \RoboFieldRevision{Go2 RL Lab MoE-CTS (mixed1.3.2resume)},
  mixed-terrain run from October 13, 2026, at 82.5k iterations.
  \RoboFieldRevision{Checkpoint released October 17, 2026}.

\bibitem[Wu et~al.(2026)Wu, Guo, Wang, Yang, Sui, Xie, Chen, Liu, and
  Lan]{suppbib-supp_go2_rllab}
Tianyang Wu, Hanwei Guo, Yuhang Wang, Junshu Yang, Xinyang Sui, Jiayi Xie,
  Xingyu Chen, Zeyang Liu, and Xuguang Lan.
\newblock {{Toward Reliable Sim-to-Real Predictability for {MoE}-based Robust
  Quadrupedal Locomotion}}.
\newblock In \emph{Robotics: Science and Systems}, 2026.

\end{thebibliography}
\end{document}